\documentclass[10pt,journal,compsoc]{IEEEtran}
\usepackage{amsmath,amssymb,amsfonts}
\usepackage[colorlinks,linkcolor=blue]{hyperref}
\usepackage{algorithmic}
\usepackage{graphicx}
\usepackage{color}
\usepackage{textcomp}
\usepackage{multirow}
\usepackage{booktabs}
\usepackage{pifont}

\ifCLASSOPTIONcompsoc
  \usepackage[nocompress]{cite}
\else
  \usepackage{cite}
\fi
\ifCLASSINFOpdf
\else
\fi
\begin{document}
\title{Structure causal models and LLMs integration in medical visual question answering}

\author{
Zibo Xu, Qiang Li, Weizhi Nie*, Weijie Wang, Anan Liu
\thanks{
This work was supported in part by the National Natural Science Foundation of China under Grants U21b2024 and 62272337. \textit{(Corresponding author: Weizhi Nie)}}
\thanks{Zibo Xu and Qiang Li are with the School of Microelectronics, Tianjin University, Tianjin 300072, China (e-mail: xzb6666@tju.edu.cn, liqiang@tju.edu.cn).}
\thanks{Weizhi Nie and Anan Liu are with the
School of Electrical and Information Engineering, Tianjin University, Tianjin 300072, China (e-mail: weizhinie@tju.edu.cn, anan0422@gmail.com). }
\thanks{Weijie Wang is with the Department of Information Engineering and Computer Science, University of Trento, Trento, Italy (e-mail: weijie.wang@unitn.it).}
\thanks{Manuscript received April 19, 2005; revised August 26, 2015.}}


\markboth{Journal of \LaTeX\ Class Files,~Vol.~14, No.~8, August~2015}%
{Shell \MakeLowercase{\textit{et al.}}: Bare Advanced Demo of IEEEtran.cls for IEEE Computer Society Journals}



\IEEEtitleabstractindextext{%
\begin{abstract}
Medical Visual Question Answering (MedVQA) aims to answer medical questions according to medical images. However, the complexity of medical data leads to confounders that are difficult to observe, so bias between images and questions is inevitable. Such cross-modal bias makes it challenging to infer medically meaningful answers. 
In this work, we propose a causal inference framework for the MedVQA task, which effectively eliminates the relative confounding effect between the image and the question to ensure the precision of the question-answering (QA) session. We are the first to introduce a novel causal graph structure that represents the interaction between visual and textual elements, explicitly capturing how different questions influence visual features. 
During optimization, we apply the mutual information to discover spurious correlations and propose a multi-variable resampling front-door adjustment method to eliminate the relative confounding effect, which aims to align features based on their true causal relevance to the question-answering task. In addition, we also introduce a prompt strategy that combines multiple prompt forms to improve the model's ability to understand complex medical data and answer accurately. Extensive experiments on three MedVQA datasets demonstrate that 1) our method significantly improves the accuracy of MedVQA, and 2) our method achieves true causal correlations in the face of complex medical data.
\end{abstract}

}

\maketitle

% is not selected <OR> if conference mode is selected - because compsoc
% conference papers position the abstract like regular (non-compsoc)
\IEEEdisplaynontitleabstractindextext

% \IEEEpeerreviewmaketitle

\section{Introduction}
\label{sec:introduction}
\IEEEPARstart{M}{edical} Visual Question Answering (MedVQA) is an important branch of the Visual Question Answering (VQA) task~\cite{Lu2023MultiscaleFE,guo2023from,tiong2022plug} in the medical field. Its main goal is to accurately answer questions according to medical images, which improves medical diagnosis and treatment efficiency. The rise of this field highlights the potential of artificial intelligence in medical diagnosis~\cite{Abacha2018NLMAI,Nie2023ChestXI,Nie2023InstrumentalVL}, providing new possibilities for improving the interpretation of multi-modal medical data and clinical decision-making. Besides, large language models (LLMs) have shown great potential in addressing the MedVQA task by enhancing the understanding of complex medical questions and generating accurate answers~\cite{llava-med,wu2023pmcllama,zhang2023pmcvqa}.
\begin{figure}[t]
  \centering
   \includegraphics[width=1\linewidth]{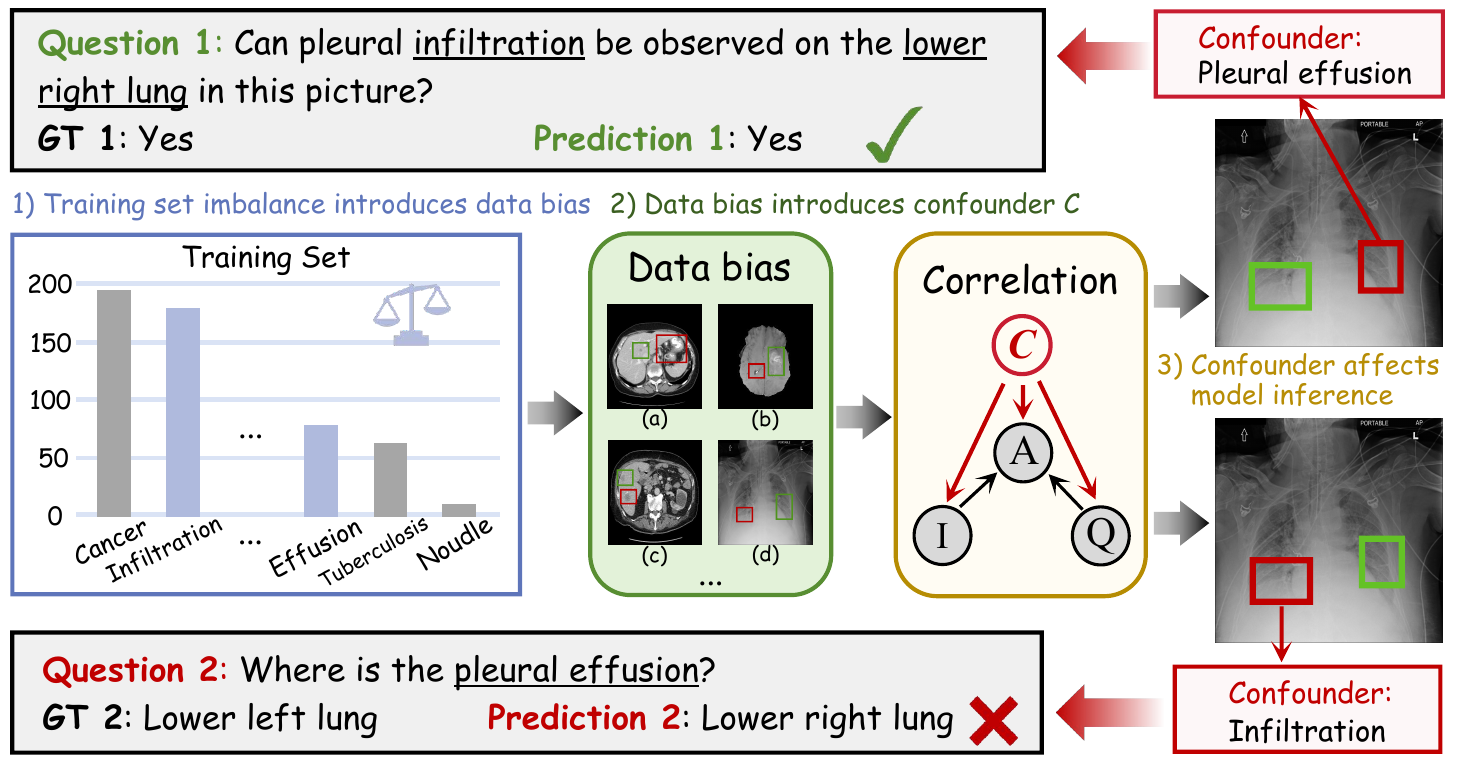}
   \caption{Illustration of how the confounder C affects model inference. The imbalanced training set leads to data bias, which can manifest in various ways, such as (a) difficulty in learning pathology, (b) statistical association, (c) relativity of confounders, and (d) co-occurrence of multiple pathologies, among others. These biases introduce confounders during inference, creating misleading causal paths. As a result, the model relies on spurious correlations rather than true causal relationships, leading to incorrect predictions.
   %Different medical questions form spurious associations with misleading areas in medical images due to data bias, leading to confounding effects. The right side of the figure illustrates common scenarios of data bias, where the purple, blue, green, and yellow boxes correspond to (a) difficulty in observing pathology, (b) statistical association, (c) relativity of confounders, and (d) co-occurrence of multiple pathologies, respectively. These confounders misalign the reasoning process by linking irrelevant features (e.g., pleural effusion) to specific question-answer pairs. As a result, the model relies on correlation instead of true causal relationships, causing incorrect predictions.
   }
   %Different medical questions form spurious associations with different misleading areas in the same medical image due to data bias. This confounding effect misleads the model's attention to spurious correlations in the training data, causing the model to make incorrect predictions.  12.13
   
   %Effect of causal intervention on VQA tasks. By separating causal features and confounders, causal intervention enables VQA models to effectively exclude false answers caused by spurious causal associations to achieve accurate answers.}
   \label{fig:fig1}
\end{figure}

%Recently, with the great power of large language models (LLMs), the MedVQA task has emerged.

Common VQA tasks~\cite{guo2023from, Lu2023MultiscaleFE} are designed to accurately respond to multi-modal inputs comprising images and queries, often focusing on natural images with diverse visual content. These tasks benefit from large-scale datasets and standardized benchmarks that allow models to learn generalizable patterns and associations effectively. However, MedVQA introduces unique challenges compared to common VQA tasks, including the complexity of medical images, which contain specialized and intricate details that require domain-specific knowledge to interpret~\cite{Zhan2020MedicalVQ,Nguyen2019OvercomingDL}. Moreover, MedVQA is particularly prone to confounding effects, such as spurious correlations between visual and textual features, which can mislead models and undermine their reliability in clinical applications.
%Common VQA tasks~\cite{guo2023from, Lu2023MultiscaleFE} are skilled at accurately responding to multi-modal inputs containing images and queries. 
%In contrast, MedVQA has certain challenges due to the model's constrained comprehension of medical images~\cite{Zhan2020MedicalVQ,Nguyen2019OvercomingDL}. 
Recent MedVQA methods~\cite{lin2023pmc,zhang2023pmcvqa,wu2023pmcllama} generally rely on large amounts of medical data for pre-training to build visual encoders or language models with medical knowledge. Some of these approaches~\cite{lin2023pmc, wu2023pmcllama} employ pre-trained encoders to build unique integrated models or introduce a projection module~\cite{zhang2023pmcvqa} to ensure alignment between medical images and questions. Although these methods have made remarkable achievements in improving the accuracy of MedVQA, they rely mainly on statistical associations and ignore the potential impact of spurious causation~\cite{Pearl2009CausalII} on the inference process. 

In the MedVQA field, it's a challenge to understand the rationale behind the answer~\cite{zhang2023pmcvqa}. Models may fail to identify confounders and spurious associations without considering the effects of multi-modal data bias~\cite{zhang2021gcastle,yuan2022auto,yx1}. In this context, excessive reliance on statistical associations may lead to wrong decisions.
When the model answers different medical questions, confounding factors often involve different areas in the image. It means that confounders corresponding to one question may be the real reason for another question, and this relativity is a difficult challenge in MedVQA tasks.

In Fig.~\ref{fig:fig1}, the medical training set is highly imbalanced, which introduces data bias and affects model inference. This data bias manifests in multiple ways: (a) difficulty in learning pathology for diseases with fewer samples, (b) statistical associations where certain regions are more frequently linked to specific diseases, (c) relative confounding, where the causal region for one question might act as a confounder for another, and (d) co-occurrence of multiple pathologies, influencing the model’s learning preferences. During training, these biases introduce confounders into the causal path, leading to spurious learning pathways and ultimately affecting model predictions.

For instance, when answering whether there is infiltration in the lower right lung, the model provides the correct answer. This is because the question provides sufficient information about both the pathology and location, and due to the larger number of infiltration samples in the training set, the model has learned the relevant features more effectively.
However, when the model is asked about the location of pleural effusion, the data bias becomes more evident. The training set contains significantly more cases of infiltration in the lower right lung compared to pleural effusion, and both pathologies share similar visual features. Due to the training data imbalance, the model is exposed to fewer pleural effusion cases, making it harder to learn its distinguishing features. As a result, during inference, the model relies more on frequently seen infiltration patterns, leading to incorrect predictions based on spurious correlations rather than true causal relationships.

The correct use of LLMs in answer generation is also a challenge~\cite{tiong2022plug}. When only a single question is provided as textual input, LLMs tend to produce a large number of irrelevant or non-answering statements as answers. This can lead to non-standard answers and a lack of practical reference value.
Above all, the issues in the MedVQA task can be summarized as the following two challenges:

1) \textbf{How to discover medical multi-modal data bias and eliminate its effects?} 
The complexity of medical data leads to biases that are difficult to observe and understand~\cite{chestcon1,chestcon2,chestcon3,yx3}. Medical images often contain intricate details that may introduce spurious associations when paired with questions, leading to confounded inference results~\cite{Pearl2009CausalII,Zhang2020CausalIF}. These biases are not only challenging to detect but are also variable, with different questions corresponding to different biases within the same medical image. This variability increases the difficulty for models to accurately interpret inputs and deduce correct answers. The presence of confounders can cause the model to focus on irrelevant or misleading associations, ultimately compromising the reliability of the diagnosis or treatment suggestions provided.

2) \textbf{How can large language models learn to generate answers with practical reference value?} 
Large Language Models (LLMs) have been introduced in MedVQA due to their exceptional capabilities in understanding and generating human-like text~\cite{llava-med,wu2023pmcllama}. However, the application of LLMs in MedVQA is not without challenges. For specific downstream tasks, it is crucial to precisely control the form of the generated text and provide ample guiding information to ensure the answers are accurate and relevant. Without such control, LLMs may generate inaccurate or invalid medical answers, which could mislead patients or medical professionals, potentially causing harm. The complexity of medical terminology and the need for precise, contextually appropriate responses further complicate this task. Ensuring that LLMs can produce answers with practical reference value involves integrating them with specialized modules that can guide their output towards medically valid and useful information.

%\xuzibo{The relativity of confounding factors needs to be fully considered in order to model causality more accurately.}

In order to solve the challenge of data bias in MedVQA, we propose a causal inference framework (CIF). The core idea of CIF is to integrate the structural causal model into the process of visual, textual processing, and answer generation. By eliminating confounding effects, CIF enables the alignment of visual and textual features based on causal relationships, ensuring that the model focuses on causally relevant regions and generates accurate, semantically aligned answers.
In this work, we identify invisible confounders in medical data by using mutual information and eliminate the effect of these relative confounders with a multi-variable resampling front-door adjustment method.
Besides, we introduce a prompt module that generates a series of prompt texts for each set of data to enhance the model's understanding of the QA pairs and images. This improves the model's ability to generate meaningful answers.

Our contributions can be summarized as:
\begin{itemize}
    \item We propose a novel multi-variable correlated causal inference framework (CIF), which first considers the relative biases caused by data bias in medical VQA. We apply the multi-variable resampling front-door adjustment, which not only considers the potential differences between the multi-modal data, but also effectively eliminates the interference of confounding factors.
    %and lack of causal inference during MedVQA, which first consider the
    %CIF effectively reduces effects of confusion and medical multimodal data bias and promotes consistency between answers, questions, and medical images.
    \item We propose a prompt strategy that integrates a prompt module into the MedVQA backbone. This strategy guides the language model to generate accurate and standardized answers by providing auxiliary information, thus enhancing the model's ability to understand and respond to questions despite the complexity of aligning different modalities.
    \item The effectiveness of our method has been strongly validated on five MedVQA datasets. These validations demonstrate the robustness and generalizability of our approach, confirming its capability to handle diverse MedVQA scenarios.

\end{itemize}

%document is a template for \LaTeX.
%You are encouraged to use it to prepare your manuscript.
%If you are reading a paper or PDF version of this document, please download the 
%\LaTeX .zip file from the IEEE Web site at \underline
%{https://www.embs.org/tmi/authors-instructions/} to prepare your manuscript.
%You can also explore using the Overleaf editor at 
%\underline
%{https://www.overleaf.com/blog/278-how-to-use-overleaf-with-}\discretionary{}{}{}\underline
%{ieee-collabratec-your-quick-guide-to-getting-started\#.}\discretionary{}{}{}\underline{xsVp6tpPkrKM9}

\section{Related Work}

\subsection{Causal Inference}
The purpose of causal inference~\cite{Pearl2009CausalII} is to pursue causal effects and eliminate false biases~\cite{Bareinboim2011ControllingSB}. In recent years, causal models have gained widespread acclaim in computer vision tasks for their superior inference abilities ~\cite{Wang2020VisualCR,Yang2020DeconfoundedIC,Zhang2020CausalIF,Li2021CausalHM,yx2}. Yue et al.~\cite{Yue2020InterventionalFL} adjust the deletion bias by backdoor adjustment and made some simple assumptions. Li et al.~\cite{Li2021CausalHM} propose a causal Markov model based on the variational autoencoder (VAE) structure to decompose disease-related variables. Yang et al.~\cite{yang2021causal} propose causal attention (CATT) based on front-door adjustment, enhancing the quality of the attention mechanism. Zhang et al.~\cite{Nie2023ChestXI,Nie2023InstrumentalVL} classify medical images through a causal perspective and utilize instrumental variables to reduce potential ambiguities in medical images. Zang et al.~\cite{Zang2023DiscoveringTR} reexamine causal effects in multi-modal data and introduces a causal prediction architecture.
However, the current causal models mainly focus on finding confounders and reducing their effects~\cite{Nie2023ChestXI}, but neglect the relativity of confounders. This can lead to poor generalization when dealing with complex cross-modal tasks.

\subsection{Medical Visual Question Answering}
The traditional MedVQA approach~\cite{Lau2018DescriptorA, Peng2018UMassAI,Zhou2018EmployingIA,Abacha2018NLMAI} aims to apply the best VQA models to the medical field; they use medical data to fine-tune VGG~\cite{Simonyan2014VeryDC} or ResNet~\cite{He2015DeepRL} for visual feature extraction. Nguyen et al.~\cite{Nguyen2019OvercomingDL} explore the use of the unsupervised Denoising Auto-Encoder (DAE)~\cite{Masci2011StackedCA} and the supervised Meta-Learning ~\cite{Vuorio2019MultimodalMM} for visual feature extraction. Based on~\cite{Nguyen2019OvercomingDL}, Zhan et al.~\cite{Zhan2020MedicalVQ} further enhance the reasoning ability of the multi-modal feature fusion module. Liu et al.~\cite{Liu2021ContrastivePA} propose a two-stage pre-training framework to tackle the challenge of data scarcity. Chen et al.~\cite{chen2022m3ae} acquire cross-modal knowledge by using random masks to reconstruct missing pixels and markers in images and text. With the popularity of LLMs, Lin et al.~\cite{lin2023pmc}, Wu et al.~\cite{wu2023pmcllama} using large amounts of medical data to pre-train models, Li et al.~\cite{llava-med} propose a model with powerful conversational capabilities and highlight that the lack of deep reasoning is a common shortcoming of existing approaches. In addition, Zhang et al.~\cite{zhang2023pmcvqa} collect an extensive data set and design a projection module to align visual and textual attributes. Huang et al.~\cite{huang2023dual} propose a dual-attention learning network for MedVQA.
However, the above methods effectively transform the model into a medical knowledge base, where the QA process involves selecting the best answers from a wide but limited pool of answers. Such methods ignore the relativity of confounders in medical data and the spurious associations caused by confounders.

\subsection{Causal Inference in MedVQA}
Recent studies have explored causal inference techniques to mitigate biases in MedVQA, aiming to improve model robustness and interpretability. Various counterfactual-based approaches have been proposed to address different types of biases. Zhan et al.~\cite{zhan2023miccai} propose a counterfactual debiasing framework to mitigate language bias in MedVQA through dual interventions.
Ye et al.~\cite{ye2024causal} propose a causal framework to mitigate modality preference bias in MedVQA by counterfactual inference. The method eliminates spurious question-answer correlations through causal path decomposition, subtracting the bias effect in scenarios where medical images are absent.
Cai et al.~\cite{cai2024counterfactual} propose a counterfactual causal-intervention strategy for MedVQA, leveraging layer-wise relevance propagation to generate interpretable saliency maps and mitigate language bias. While these methods mitigate biases through counterfactual inference, they mainly focus on direct associations. However, MedVQA involves complex interactions where indirect causal effects also contribute to biases.

%Use one space after periods and colons. Hyphenate complex modifiers: 
%``zero-field-cooled magnetization.'' Avoid dangling participles, such as, 
%``Using \eqref{eq}, the potential was calculated.'' It is not clear who or what 
%used \eqref{eq}. Write instead, ``The potential was calculated by using \eqref{eq},'' or 
%``Using \eqref{eq}, we calculated the potential.''

%Use a zero before decimal points: ``0.25,'' not ``.25.'' Use 
%``cm$^{3}$,'' not ``cc.'' Indicate sample dimensions as ``0.1 cm 
%$\times $ 0.2 cm,'' not ``0.1 $\times $ 0.2 cm$^{2}$.'' The 
%abbreviation for ``seconds'' is ``s,'' not ``sec.'' Use %5``Wb/m$^{2}$'' or ``webers per square meter,'' not 
%``webers/m$^{2}$.'' When expressing a range of values, write ``7 to 
%9'' or ``7--9,'' not ``7$\sim $9.''

%A parenthetical statement at the end of a sentence is punctuated outside of 
%the closing parenthesis (like this). (A parenthetical sentence is punctuated 
%within the parentheses.) In American English, periods and commas are located within 
%quotation marks, like ``this period.'' Other punctuation is placed ``outside''! 
%Avoid contractions; for example, write ``do not'' instead of ``don't.'' The 
%serial comma is preferred: ``A, B, and C'' instead of ``A, B and C.''

%If you wish, you may write in the first person singular or plural form using
%the active voice (``I observed that $\ldots$'' or ``We observed that $\ldots$'' 
%instead of ``It was observed that $\ldots$''). Remember to check spelling. If 
%your native language is not English, please have a native English-speaking 
%colleague to carefully proofread your paper.

\section{Methodology}
\subsection{Causal Analysis of MedVQA}
\begin{figure}
  \centering
   \includegraphics[width=\linewidth]{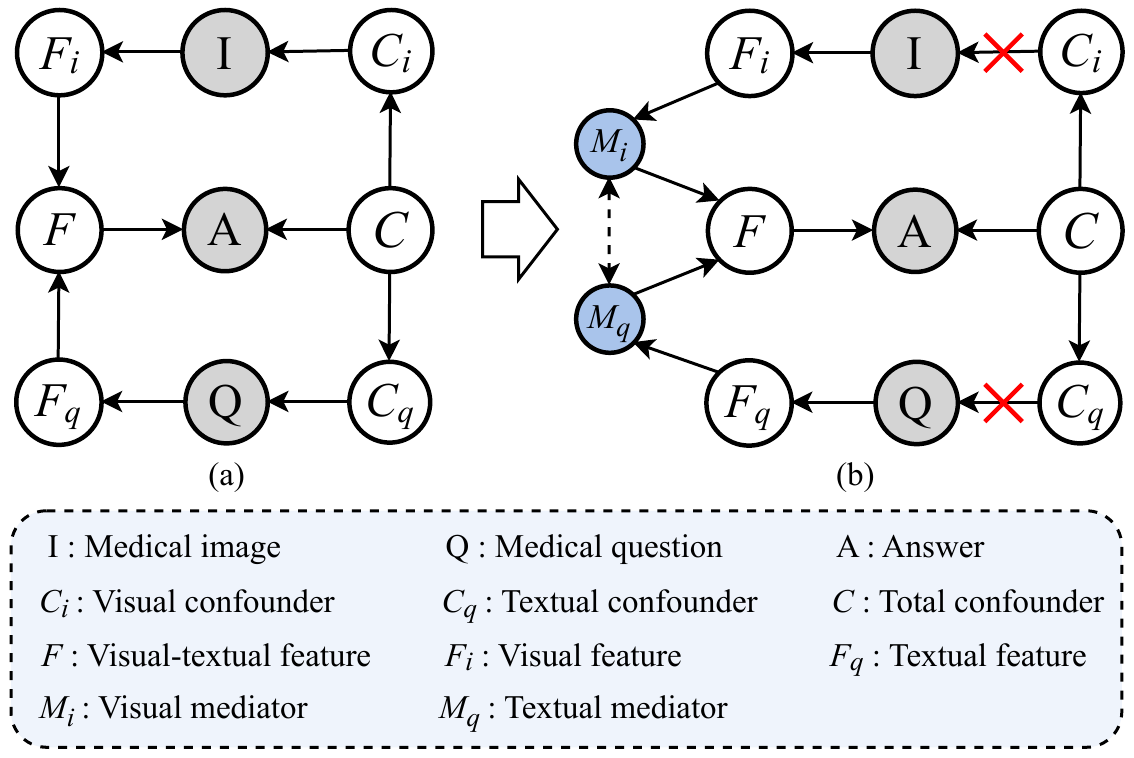}
   \caption{Causal directed acyclic graph of MedVQA, gray variables are observed data, and blue variables are mediators. \textbf{(a)} Important variables in MedVQA and their associations.
   %We use the do operator to cut off the path to $\left \{I,Q\right \}$.
   \textbf{(b)} We apply the front-door adjustment by introducing two mediators to deal with the invisible confounders. This reveals the true causal relationship between $\left \{I,Q\right \}$ and $A$. The dashed line between the two mediators reflects the relative confounders considering the interaction between the two modalities.}
   \label{fig:cau}
\end{figure}

We use the structural causal model~\cite{Pearl2009CausalII,Liu2022CrossModalCR} to represent the causal relationships in the MedVQA task. Each node represents a key variable, and the connecting lines represent the causal relationship between the variables, as shown in the causal graph Fig. \ref{fig:cau}(a).

$I\gets C_{i}\gets C\to A$ and $Q\gets C_{q}\gets C\to A$ are the backdoor paths of $\left \{I,Q\right \}\to F\to A$ respectively, where total confounders $C$ consist of visual confounders $C_{i}$ and textual confounders $C_{q}$. $C$ arises from biases in medical multi-modal data, leading to spurious associations between $\left \{I,Q\right \}$ and $A$. For example, a spurious, strong correlation may appear between a medical question and a pathology in an image or between two pathologies in an image. When answering according to $I$ and $Q$, if the model only learns the statistical association $P(A|I,Q)$, it will ignore the influence of spurious associations raised by confounders $C$, leading to generating incorrect answers.

$I\to F_{i}\to F\gets F_{q}\gets Q$ represents the extraction and fusion process of medical image features and textual features. In this process, $I$ and $Q$ are converted into features $F_{i}$ and $F_{q}$ by feature extraction modules. Subsequently, $F_{i}$ and $F_{q}$ fuse into the multi-modal joint feature $F$, and the efficient capture and integration of $F$ form the basis of true causality. However, due to the existence of the backdoor paths, $F$ extracted by learning the statistical association $P(A|I,Q)$ is bound to be affected by confounders C, resulting in some confounding features and false causal associations in $F_{i}$ and $F_{q}$.

$F\to A$ represents a direct connection between the fusion feature and the answer. Ideally, this connection should be independent of the variable $C$, and the path should be the only source from which the answer is generated.

\begin{figure*}[th]
 \centering
	\includegraphics[width=\textwidth]{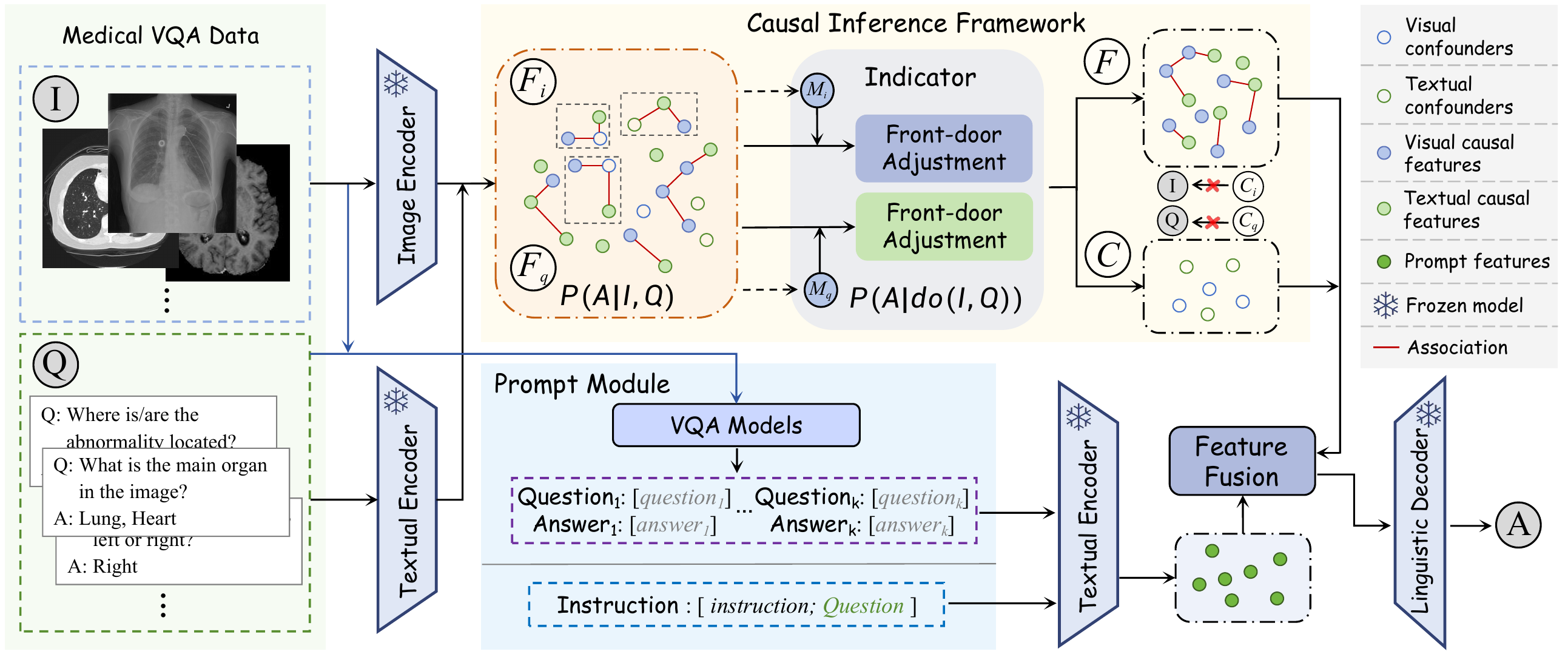}
	\caption{Overview of our method. There are causal features, confounding features, and their spurious associations in the feature space. We introduce an indicator module to obtain true causal features in MedVQA data, where we use mutual information to adjust the factors that may lead to spurious associations when learning mediators. For the features of two modalities, we use a multi-variable resampling front-door adjustment method to separate causation and confounding factors.
 %By learning the information of causal features and confounding features, the model strengthens the influence of real causal associations, thus significantly improving the question-answering ability. 
 The prompt module receives a variety of inputs, including instruction, questions, and a new set of QA pairs generated from each original data, which are then fed into the pre-trained linguistic decoder as an overall prompt for answering.}
    \label{net}
\end{figure*}

Taking the example in Fig. \ref{fig:fig1}, where $P(c=$ ``infiltration'' $|I=$ ``abnormal area''$,Q=$ ``right''$)$ is much larger than $P(c=$ ``effusion'' $|I=$ ``abnormal area''$,Q=$ ``right''$)$, so $P(A|I,Q,c=$ ``infiltration''$)$ plays a more important role than $P(A|I,Q,c=$ ``effusion''$)$ when calculating $P(A|I,Q)$. This leads the model to over-rely on the spurious correlation between the ``right'' lung and ``infiltration'' rather than correctly distinguishing between pathologies. The causal inference intervenes by breaking these spurious correlations, ensuring that 
$P(A|do(I,Q))$ is computed based on causal relationships rather than statistical associations. 
To achieve this, we model the conditional probability by explicitly integrating causal paths within the network design. Specifically, the structural causal model provides a mechanism to estimate and adjust for the influence of confounders, aligning $P(A|do(I,Q))$ with its true causal components. 

In subsequent sections, we demonstrate how the probability is implemented within the network through specific modules, where the causal adjustment process is implemented as part of the alignment of multi-modal features. 
%This design ensures that the network adheres to the causal structure of the task, effectively eliminating the confounding effects while maintaining robust alignment between visual and textual features.

\subsection{Causal Inference Techniques in MedVQA}

Ideally, we can employ causal analysis and causal inference to intervene with $\left \{I,Q\right \}$, ensuring that the model avoids erroneously capturing spurious correlations $\left \{I,Q\right \}\gets C\to A$ introduced by confounders $C$, and instead accurately captures real causal structure $\left \{I,Q\right \}\to A$~\cite{Liu2022CrossModalCR}.

\subsubsection{Front-door adjustment} 
Confounders in medical multi-modal data are often complex and unobserved. In this case, front-door adjustment is a feasible method of causal intervention~\cite{Pearl2009CausalII}.
The core idea of causal intervention is to cut off the backdoor path to $\left \{I,Q\right \}$ by $do(I,Q)$, as shown in Fig. \ref{fig:cau}(b). However, since $C_{i}$ and $C_{q}$ are often unobserved, we introduce two mediators, $m_i$ and $m_q$ to implement the process of removing confounders. In MedVQA tasks, the data of the two modalities are often closely related, so $m_i$ and $m_q$ are not actually independent. We combine Fig. \ref{fig:cau}(b) with the front-door adjustment:
\begin{align}
&P(A|do(I,Q))  \notag \\
&= \sum_{m} P(A|do(M=m)) P(M=m|do(I,Q)) \notag \\
&= \sum_{m\in \mathcal{M}}\sum_{i\in I}\sum_{q\in Q} P(A|m,i,q)P(i,q)P(m|I,Q)\notag \\
&= \sum_{m\in \mathcal{M}}P(m|I,Q)\sum_{i\in I}\sum_{q\in Q} P(A|m,i,q)P(i,q),
\label{eq:fd44}
\end{align}
where $m$=$\left \{m_{i}^{1},\cdots,m_{i}^{k},m_{q}^{1},\cdots,m_{q}^{n}\right \}$ is the set of mediators, $k$ and $n$ are related to images and questions, respectively. $i$ and $q$ are approximately sampled from $\left \{I,Q\right \}$ at the feature level~\cite{pmlr-v37-xuc15}. $P(A|m,i,q)$ represents the combined influence of preprocessed medical data and mediators on answers. 
%See \textbf{Appendix} for a derivation of MedVQA's front-door adjustments. 

Since the answer generation part of the MedVQA task is essentially to select the best answer from a large corpus, $P(A|m,i,q)$ can be approximated as a network g(·) followed by softmax~\cite{yang2021causal}:
\begin{equation}
\begin{aligned}
P(A|m,i,q)=\text{Softmax}(g(m,i,q)).
\end{aligned}
\end{equation}
Theoretically, Eq.~\ref{eq:fd44} requires sampling all medical data and corresponding mediators. To simplify this process, we use the Normalized Weighted Geometric Mean (NWGM)~\cite{pmlr-v37-xuc15} to further approximate:
\begin{equation}
\begin{aligned}
&WGM(f(x))=\prod_{x}f(x)^{P(x)}=\prod_{x}exp[g(x)]^{P(x)}\\
&=exp[\sum_{x}g(x)P(x)]=exp[\mathbb{E}_x[g(x)]]\approx \mathbb{E}_x[f(x)],\\
&N(f(x))=\frac{\prod_{x}exp(g(x))^{p(x)}}{\sum_{j}\prod_{x}exp(g(x))^{p(x)}}
=\text{Softmax}({\mathbb{E}_x[g(x)]}).
\label{nwgm}
\end{aligned}
\end{equation}
where function $W$ denotes Weighted Geometric Mean and $N$ denotes NWGM, $f(x)=exp[g(x)]$. Then the sampling process can be expressed as:

\begin{equation}
\begin{aligned}
&\sum _{i\in I}\sum _{q\in Q}P(A|m,i,q)P(i,q)=\mathbb{E} _{i,q}\text{Softmax}(g(m,i,q)) \\
&=\text{Softmax}(g(m,\mathbb{E} _{i,q}(i,q)))=\text{Softmax}(g(m,\boldsymbol{i},\boldsymbol{q})),
  \label{eq:fd30}
\end{aligned}
\end{equation}
where $\mathbb{E}_{i,q}$ denotes the expectation function, $\boldsymbol{i}$, $\boldsymbol{q}$ denote the estimations of $i$, $q$. Then $P(A|do(I,Q))$ can be obtained as:
\begin{equation}
\begin{aligned}
&P(A|do(I,Q))=\sum_{m\in \mathcal{M}_i}\text{Softmax}(g(m,\boldsymbol{i},\boldsymbol{q}))P(m|I,Q)\\
&=\text{Softmax}[G(\boldsymbol{m},\boldsymbol{i},\boldsymbol{q})]=\text{Softmax}[G(\boldsymbol{m_{i}},\boldsymbol{m_{q}},\boldsymbol{i},\boldsymbol{q})],
  \label{eq:fd301}
\end{aligned}
\end{equation}
where $G($·$)$ represents the overall causal inference network, and $\boldsymbol{m}$ denotes the estimations of $m$. 

We approximate causal intervention $P(A|do(I,Q))$ as the process of combining medical visual and textual features with their corresponding mediators. We introduce mutual information to quantify the correlation between images and questions, ensuring that the sampling processes of $\boldsymbol{m_{i}}$, $\boldsymbol{m_{q}}$ are based on the true correlations between the two modalities. 
$\boldsymbol{i}$ and $\boldsymbol{q}$ are essentially extracted features $f_{i}$ and $f_{q}$, presenting in $F_{i}$ and $F_{q}$ in the causal graph Fig. \ref{fig:cau}. $\boldsymbol{m_{i}}$, $\boldsymbol{m_{q}}$ are calculated by mediator integrators. The detailed deconfounding process is shown in Fig. \ref{fig:ind}.
%\begin{equation}%
%\begin{aligned}
%& \boldsymbol{m_{i}} = \sum_{m_{i}\in \mathcal{M}_i} P(m_{i}|f_{1}(i,q))m_{i}, \boldsymbol{m_{q}} = \sum_{m_{q}\in \mathcal{M}_q} P(m_{q}|f_{2}(i,q))m_{q},
%\end{aligned}
%\label{eq:iqmm}
%\end{equation}
%\boldsymbol{i} = \sum_{i\in I} P(i|\Phi_{ie}(i))i,\quad \boldsymbol{q} = \sum_{q\in Q} P(q|\Phi_{qe}(q))q,\quad
%where $\Phi_{ie}$, $\Phi_{qe}$ are feature extraction functions, and $f_{1}$, $f_{2}$ are feature fusion networks based on Multi-Head Attention (MHA).  

Through the steps above, we have expressed the causal effect as a combination of image and question features. However, to address the potential confounders in both image and text, we introduce two specialized sub-networks: the visual de-confounding network and the textual de-confounding network: 
\begin{align}
&P(A|do(I, Q)) \notag \\  &\propto \text{Softmax}\left(\sum_{m_i \in \mathcal{M}_i} \sum_{m_q \in \mathcal{M}_q} Y(m_i, f_{i}, f_{q}) \cdot H(m_q, f_{i}, f_{q})\right),
\label{eq:YH}
\end{align}
where the functions \( Y(m_i, f_i, f_q) \) and \( H(m_q, f_i, f_q) \) represent the interactions between image and question features. In Eq.~\ref{eq:YH}, the interactions between features is jointly modeled through the de-confounding networks \( Y \) and \( H \), effectively removing confounders in both image and question, and enabling an accurate estimation of the causal effect.

In the next section, we will provide a detailed explanation of how to implement these two de-confounding networks, particularly how they utilize attention mechanisms and feature interactions to eliminate confounders, improving the robustness and performance of the model.

\subsubsection{Implement method}
%For medical images, we employ a pre-trained visual feature extraction model~\cite{lin2023pmc} to capture visual details, including pathology, organ structures, and imaging techniques~\cite{Nie2023ChestXI}
Given the medical image $i\in \mathbb{R}^{H\times W\times 3}$ as input, we use the pre-trained image encoder Clip ViT-B/16~\cite{clip} to extract the feature $f_{i}\in \mathbb{R}^{h\times w\times d} $, where $H\times W$ and $h\times w$ represent the height and width of the medical image and the feature map, respectively, and $d$ represents the hidden dimension of the network.
\begin{figure*}

  \centering
   \includegraphics[width=\linewidth]{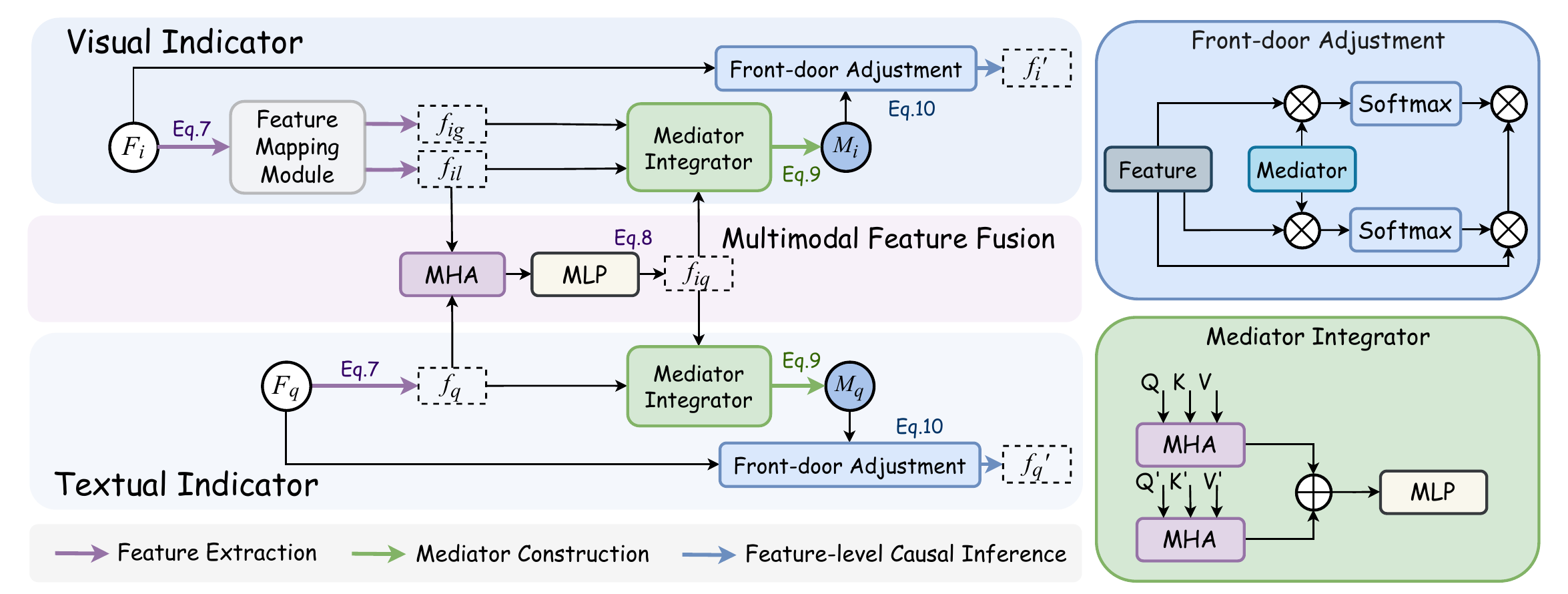}
   \caption{Overview of indicator structure. The indicator structure consists of a visual indicator, a text indicator, and a multi-modal feature fusion module. This structure generates two mediators, which are combined with $F_{i}$ and $F_{q}$ to obtain true causal features by front-door adjustment.}
   \label{fig:ind}
\end{figure*}
Medical questions involve semantic content such as image types, organs, and pathology, while answers are often short, typically consisting of only one to four words~\cite{Lau2018DescriptorA,40cb06d16fd1450ea39bfd13d43e9c9f,He2020PathVQA3Q}. 
We use a hierarchical semantic parser~\cite{Liu2022CrossModalCR} to transform medical questions into image-based markers, extract semantic associations from the medical knowledge base, and decompose the questions into different forms. The parsed outputs are then processed by the frozen LLaMA encoder, which encodes these structured semantic representations into high-dimensional embeddings. This hierarchical parsing and encoding pipeline ensures that the questions are semantically enriched and properly aligned with the model’s visual features, facilitating causal reasoning and accurate cross-modal alignment.
%We use a hierarchical semantic parser~\cite{Liu2022CrossModalCR} to transform medical questions into image-based markers, extract semantic associations from the medical knowledge base, and decompose the questions into different forms. In the training process, in order to capture the contextual semantic relationship of complex questions, we use a pre-trained language encoder~\cite{wu2023pmcllama} to encode the question types, semantics, key points, and answers after word segmentation. 
The two feature extraction processes are as follows:
\begin{equation}
\begin{aligned}
f_{i} &= \Phi_{ie}(i, \Theta_i), & f_{q} &= \Phi_{qe}(q, \Theta_q), \\
\end{aligned}
\end{equation}
where $\Phi_{ie}$, $\Phi_{qe}$ represent the pre-trained image encoder and the textual encoder, and $\Theta_{i}$ and $\Theta_{q}$ represent their corresponding parameters.
 
The Feature Mapping Module (FMM) is developed to extract global features \( f_{ig} \) and local features \( f_{il} \) from medical images, enabling complementary representations for downstream tasks. For global sampling, FMM employs a Down-Sampling Transformer block that aggregates visual tokens to retain the overall structure and spatial layout of the image. This captures macroscopic details, such as the relative positioning and contours of organs, providing crucial contextual information for reasoning. For local sampling, inspired by attention mechanisms, the FMM leverages accumulated attention maps to identify and extract the top \( k \) tokens that correspond to regions of interest, such as lesions or abnormalities. For each attention head, k=8 tokens are selected, providing granular and contextually significant details that align with the limited image diversity. This ensures that fine-grained details essential for precise question answering are effectively captured. The combination of these two sampling strategies enhances the model's capacity to align global context with local precision, which is particularly vital for the complex reasoning required in MedVQA tasks.

In MedVQA, the model needs to establish the correct connection between the medical image and the question. In this case, we use a feature fusion module to generate the cross-modal interaction features $f_{iq}$:
\begin{equation} 
\begin{aligned} &f_{iq}=MLP(MHA(f_{il},f_{q},f_{q})), \end{aligned} 
\end{equation}
MHA and multi-layer perceptron (MLP) constitute feature fusion modules. Here, 
$f_{il}$ represents the local visual feature extracted from the image, while $f_{q}$
is the textual feature derived from the question. This structure ensures that local image features remain central, while question features are incorporated to guide attention and align the semantic meaning across modalities.
According to Fig. \ref{fig:ind}, in the indicator structure, we use front-door adjustment for visual features and linguistic features, respectively. Considering the interaction between the two modalities and the relativity of confounders, we use mutual information and multivariate resampling methods to sample the two mediators:
\begin{equation} \begin{aligned} &\boldsymbol{m_{i}}=MLP(MHA(f_{il},f_{ig},f_{ig}),MHA(f_{il},f_{iq},f_{iq})),\\ &\boldsymbol{m_{q}}=MLP(MHA(f_{q},f_{q},f_{q}),MHA(f_{iq},f_{q},f_{q})), \label{mha} \end{aligned} \end{equation}
where $\boldsymbol{m_{i}}$ and $\boldsymbol{m_{q}}$ are the visual and textual mediators, respectively.
The rationale for these designs is as follows. For $\boldsymbol{m_{i}}$, $MHA(f_{il},f_{ig},f_{ig})$ captures the relationship between local features $f_{il}$ and global features $f_{ig}$, while $MHA(f_{il},f_{iq},f_{iq})$ integrates question-specific focus into the visual feature representation. This combination enables the model to account for both overall structural information and localized details relevant to the question. For $\boldsymbol{m_{q}}$,
$MHA(f_{q},f_{q},f_{q})$ emphasizes linguistic coherence within the question text, while $MHA(f_{iq},f_{q},f_{q})$ incorporates cross-modal semantic alignment between textual and visual features. These mediator designs are carefully chosen to balance intra-modal dependencies and cross-modal interactions, providing robust causal feature adjustments.

The extracted multi-modal feature $f_{iq}$ contributes to the interaction between the two modalities when extracting
$\boldsymbol{m_{i}}$ and $\boldsymbol{m_{q}}$ in Eq.~\ref{mha}, and helps the model focus on the confounding areas specified by the question. 
$\boldsymbol{m_{i}}$ reveals the relationship between the local features, global features, and fusion features of medical images. 
$\boldsymbol{m_{q}}$ reveals the associations between different parts of the medical question text and the semantic connections between fusion features and textual features.

Areas of interference in the same medical image vary with different questions. This means that word features affect areas of concern for visual features through cross-attention. Considering this relativity of confounders, we resample the mediators to ensure that we can discover and adjust for the proprietary confounder for each question. This dynamic adjustment ensures that
$\boldsymbol{m_{i}}$ and $\boldsymbol{m_{q}}$ capture the essential causal relationships between modalities while effectively mitigating confounding effects.
By introducing the front-door adjustment module, the entire process of causal inference and the processes of deconfounding the causal and confounding factors of multi-modal medical data can be represented as:
\begin{equation}
\begin{aligned}
P(A|do(I,Q))=\text{Softmax}&[\boldsymbol{g}(Y(f_{i},\boldsymbol{m_{i}}),H(f_{q},\boldsymbol{m_{q}}))],\\
f_{i}^{'}=Y(f_{i},\boldsymbol{m_{i}}),&f_{q}^{'}=H(f_{q},\boldsymbol{m_{q}}),
\end{aligned} 
\end{equation} 
where $\boldsymbol{g}($·$)$ represents a feature fusion network, $Y$, $H$ represent the visual front-door adjustment and the textual front-door adjustment, $f_{i}^{'}$,  $f_{q}^{'}$ are visual causal features, and textual causal features, respectively.

\subsection{Prompt Module }
In MedVQA, random outputs generated by LLMs may result in only a small fraction of the content matching the ground truth. In order to obtain medically meaningful answers, we design a Prompt Module (PM). PM is designed for MedVQA, which poses greater challenges compared to natural images due to the presence of complex structures. Our approach involves leveraging pre-trained visual and linguistic backbones to effectively generate QA pairs, and we design different templates for reference of LLMs. In essence, we position PM as a pivotal component complementing CIF, enhancing the effectiveness of LLMs in generating answers by incorporating causal features. 

The Prompt Module (PM) is seamlessly integrated into the MedVQA framework to enhance the reasoning capabilities of LLMs by embedding structured prompts into the QA stage. Specifically, we employ two distinct types of instructions to serve different purposes.
For QA pairs generation, \( \text{Instructions}_\text{gen} \) guide the generation of QA pairs based on medical images, ensuring alignment with clinical reasoning. These predefined instructions are critical for creating meaningful QA pairs and are detailed in Table \ref{prompts}.  
For task introduction, \( \text{Instructions}_\text{task} \) provide context to the LLM for the final output generation. For example, "You are a doctor, please answer the question based on the medical image and QA pairs." While \( \text{Instructions}_\text{task} \) is not essential to model performance, it helps standardize interactions with the LLM by briefly introducing the task requirements.  

Besides, we precede the input question with a series of generated QA pairs $\left \{Q_{n};[Question],A_{n}:[Answer]\right \}$ for each set of Image-QA pairs, acting as exemplars to guide the model. These prompts serve as caption-like guidance, helping the model focus on causal patterns rather than relying solely on textual cues. The specific template is shown in Table \ref{prompts}.  

The generation process for the prompts is straightforward and efficient. Using the pre-trained visual and linguistic backbones (CLIP ViT-B/16 and LLaMA), the model directly infers QA pairs from each original data point in the dataset. The generated QA pairs take the form, representing meaningful examples that align with the underlying medical context. The prompt structure integrates these generated QA pairs along with task instructions and the current question, formatted as $\{Instructions_{task}$,$[QA]_{n}$,$Question$:$[Question]$,$A:$$\}$, \( n \in [1, 3]\). For the prompt text, we use the same feature extraction method as for medical questions.

\iffalse
\begin{table}[h]
\centering
\caption{Templates for different types of questions, used to generate QA pairs.}
\scalebox{0.9}{\begin{tabular}{ccc}
\toprule
\multicolumn{1}{l}{\textbf{Type}} & \multicolumn{1}{l}{\textbf{Type}} & \textbf{Template of QA pairs} \\
\midrule
\textit{QA Pairs}& & \\
\multirow{3}{*}{Open} & What & What [$\mathit{organ(s)/disease(s)}$] is/are in the image? \\
                      & Which & Which [$\mathit{disease(s)/organ(s)}$] is/are in the image?\\
                      & Where & Where is/are the [$\mathit{abnormality/organ(s)}$] located?\\
\midrule
\multirow{3}{*}{Closed} & Is & Is this a/an [$\mathit{imaging\, techniques/organ}$]?\\
                        & Does & Does the image contain [$\mathit{organ/abnormality}$]?\\
                        & Which & Which side is [$\mathit{organ/abnormality}$] in the image?\\
\bottomrule
\textit{Instructions}& & \\
\end{tabular}
}
\label{prompts}
\end{table}
\fi

\begin{table}[h]
\centering
\caption{Templates for different types of questions used to generate QA pairs.}
\scalebox{0.9}{
\begin{tabular}{cc}
\toprule
\multicolumn{1}{c}{\textbf{Type}} & \multicolumn{1}{c}{\textbf{Template of QA pairs}} \\
\midrule
\multicolumn{1}{l}{\textit{QA Pairs}}& \\
\hspace{1em}What & What [$\mathit{organ(s)/disease(s)}$] is/are in the image? \\
\hspace{1em}Which & Which [$\mathit{disease(s)/organ(s)}$] is/are in the image? \\
\hspace{1em}Where & Where is/are the [$\mathit{abnormality/organ(s)}$] located? \\
\hspace{1em}Is & Is this a/an [$\mathit{imaging\, techniques/organ}$]? \\
\hspace{1em}Does & Does the image contain [$\mathit{organ/abnormality}$]? \\
\hspace{1em}Which & Which side is [$\mathit{organ/abnormality}$] in the image? \\
\midrule
\multirow{5}{*}{\textit{Instructions}} & Describe the following medical image in detail. \\
& Answer the question based on the medical image. \\
& Provide a detailed analysis of the medical image. \\
& Analyze the medical image and describe any abnormalities. \\
& Based on the image, provide a differential diagnosis. \\
\bottomrule
\end{tabular}
}
\label{prompts}
\end{table}

\subsection{Training Strategy}
%The answer-generation process can be expressed as:
%where $\boldsymbol{\Phi_{d}}$ represents the pre-trained linguistic decoder and $f_{p}$ represents the features of the total prompt text.
For closed questions and open-ended questions, we design different loss functions, for closed questions, we apply supervised loss in a cross-entropy format:
\begin{equation}
\begin{aligned}
&\mathcal{L}_{c}=-\frac{1}{\left | \textbf{T} \right | }\sum_{\boldsymbol{t}\in \textbf{T}}\boldsymbol{a_{t}}^{\top}log(\textbf{V}(i,q)), 
  \label{eq:lossc}
\end{aligned}
\end{equation}
where $\boldsymbol{a_{t}}$ represents the ground-truth answer, $\textbf{T}$ is the training data, $\textbf{V}($·$)$ denotes our proposed framework.

For open-ended questions, the loss function is:
\begin{equation}
\begin{aligned}
\mathcal{L}_{o}=-\sum_{\boldsymbol{n}=1}^{\textbf{N}}log\textbf{p}(\boldsymbol{a_{n}}|i,q,\boldsymbol{p_{t}},\boldsymbol{a_{1:n-1}};\Theta_{o} ),
  \label{eq:losso}
\end{aligned}
\end{equation}
where $\textbf{N}$ is the length of the ground-truth, $\boldsymbol{a_{1:n-1}}$ denotes previous tokens, $\Theta_{o}$ denotes the parameter of the model, and $\boldsymbol{p_{t}}$ is the prompt text.
Eq.~\ref{eq:losso} allows the model to generate answers based on multi-modal inputs and partial outputs, which is more efficient than categorizing in candidate answer sets.
%$\textbf{p}(\boldsymbol{a_{o}}|\boldsymbol{i},\boldsymbol{q},\boldsymbol{p_{t}},\boldsymbol{a_{1:n-1}};\Theta_{o} )$ represents the probability of generating the n-th token in the answer sequence given $\boldsymbol{i}$, $\boldsymbol{q}$, $\boldsymbol{p_{t}}$ and $\boldsymbol{a_{1:n-1}}$, 
%(\theta)
To make the predictions of original features and causal features consistent and highlight the true causal effect, we design a loss function:
\begin{align}
\begin{split}
\mathcal{L}_{cau}=\textbf{KL}(\textbf{V}(f_{i}^{'},f^{'}_{q}),\textbf{V}(f_{i},f_{q})),
  \label{eq:lcaud}
\end{split}
\end{align}
where $\textbf{KL}$ is the KL-divergence.
%$d$ in ~\ref{ddd} represents samples in a updatable feature dictionary $D$ which approximates confounder $C$.
Finally, the overall learning objectives for two forms of questions are:
\begin{align}
\begin{split}
\mathcal{L}_{closed}=\mathcal{L}_{c}+\mathcal{L}_{cau},
\mathcal{L}_{open}=\mathcal{L}_{o}+\mathcal{L}_{cau}.
  \label{eq:lossn}
\end{split}
\end{align}
For datasets such as ProbMed and PMC-VQA, where questions are not explicitly categorized into closed-ended and open-ended forms, we adopt the loss function designed for open-ended questions.
This choice is motivated by the fact that answers in these datasets tend to be more diverse and detailed, making the open-ended loss function more suitable.

\begin{table*}[h]
\centering
\caption{Comparison of accuracy with state-of-the-art methods on three MedVQA datasets. Questions are divided into open and closed questions. Also, ``Overall'' indicates the overall accuracy over each entire dataset. The second-best results are underlined.}
%\scalebox{0.7}
\resizebox{0.8\textwidth}{!}
{
\begin{tabular}{lccccccccc}
\toprule
\multicolumn{1}{c}{\multirow{2}{*}{\multirowsetup{\centering}\textbf{Method}}}      & \multicolumn{3}{c}{\textbf{VQA-RAD}
} & \multicolumn{3}{c}{\textbf{SLAKE}} & \multicolumn{3}{c}{\textbf{PathVQA}} \\
\cmidrule(lr){2-4} \cmidrule(lr){5-7} \cmidrule(lr){8-10}
            & \multicolumn{1}{c}{Open}  & \multicolumn{1}{c}{Closed} & \multicolumn{1}{c}{Overall} & Open  & Closed & Overall & Open & Closed & Overall \\
\midrule
MEVF-BAN~\cite{Nguyen2019OvercomingDL}    & 49.2  & 77.2  & 66.1 & 77.8  & 79.8  & 78.6 & 8.1 & 81.4 & 44.8 \\
CPRD-BAN~\cite{Liu2021ContrastivePA}    & 52.5  & 77.9  & 67.8 & 79.5  & 83.4  & 81.1 & - & - & - \\
M3AE~\cite{chen2022m3ae}        & 67.2  & 83.5  & 77.0 & 80.3  & 87.8  & 83.3 & - & - & - \\
PMC-CLIP~\cite{lin2023pmc}    & 67.0  & 84.0  & 77.6 & 81.9  & 88.0  & 84.3 & - & - & - \\
CLIP-ViT~\cite{Sonsbeek2023OpenEndedMV}   & -  & -  & - & 84.3  & 82.1  & 83.3 & 40.0 & 87.0 & 63.6 \\
M2I2~\cite{M2I2}    & 66.5  & 83.5  & 76.8 & 74.7  & \underline{91.1}  & 81.2 & 36.3 & 88.0 & 62.2 \\
LLaVA-Med~\cite{llava-med}    & 64.4  & 82.0  & 74.9 & 84.7  & 83.2  & 84.1 & 38.9 & \textbf{91.7} & \underline{65.3} \\
MedVInT-TE~\cite{zhang2023pmcvqa}  & 69.3  & 84.2  & 78.2 & 88.2  & 87.7  & 88.0 & - & - & - \\
MedVInT-TD~\cite{zhang2023pmcvqa}  & 73.7  & 86.8  & 81.6 & 84.5  & 86.3  & 85.2 & - & - & - \\
\midrule
\textbf{Ours (7B)}        & \underline{74.3}  & \underline{87.1} & \underline{82.0} & \underline{90.1}  & 90.4 & \underline{90.2} & \underline{40.4} & 87.9 & 64.2 \\
\textbf{Ours (13B)}        & \textbf{76.0}  & \textbf{87.9} & \textbf{83.1} & \textbf{90.5}  & \textbf{91.8} & \textbf{91.0} & \textbf{41.4} & \underline{91.5} & \textbf{66.5} \\
\bottomrule
\end{tabular}}
\label{table:your_table_label}
\end{table*}
\section{Experiment}
\label{sec:Experiment}

\subsection{Datasets}

\noindent\textbf{VQA-RAD}~\cite{Lau2018DescriptorA} is a dataset specifically designed for radiology, consisting of 315 images and 3,515 questions with 517 possible answers. The dataset includes 3,064 tasks for training and 451 tasks for testing. All questions are concise, typically ranging from 5 to 7 words, while the answers are even shorter, averaging about 1.6 words each.

\noindent\textbf{SLAKE}~\cite{40cb06d16fd1450ea39bfd13d43e9c9f} is an English-Chinese bilingual MedVQA dataset consisting of 642 images and 14k QA pairs, covering 12 diseases and 39 systemic organs. Diseases mainly include tumors and chest diseases, and the images mainly include the head, chest, abdomen, pelvic cavity, and other body parts. 
We follow the original dataset splitting,
where 4,919 tasks about 450 images are used for training,
1,053 tasks about 96 images for validation, and 1,061 tasks
about 96 images for testing.
%All of the images are divided into 282 CT, 181 MRI, and 179 X-ray images according to image types, among which all CT and MRI are axial single-row. The number of images for each body part is set according to the complexity of the body part.

\noindent\textbf{PathVQA}~\cite{He2020PathVQA3Q} is a pathological image dataset containing 4998 pathological images, and 32,799 QA pairs. Each image is accompanied by multiple questions covering various aspects such as location, shape, color, appearance, etc.
In the official dataset split, the training set, validation set and test set contain 19,755, 6,279
and 6,761 QA pairs, respectively.

\noindent\textbf{PMC-VQA}~\cite{pmcvqa} is a large-scale MedVQA dataset of 227,000 QA pairs that cover 149,000 images, covering multiple modes of medical imaging and types of diseases. It was constructed to support and facilitate the development of MedVQA models, especially in generative QA tasks, capable of addressing diverse questions that arise in clinical practice. All questions are multiple-choice. In the official dataset split, the training set and test set 82
\% and 18\%.

\noindent\textbf{ProbMed}~\cite{probme} is a probe evaluation dataset for medical diagnosis, containing 6,303 images and 57,132 QA pairs covering a wide range of image modes and organs. The dataset requires the model to reason across multiple diagnostic dimensions, including modal recognition, organ recognition, clinical findings, anomalies, and location reasoning.
Since the dataset was not explicitly divided officially, we randomly divided it into the training and test sets at a ratio of 4:1.

\subsection{Implementation details}
%We choose the LLaMA~\cite{wu2023pmcllama} as the language backbone of our framework, and CLIP as the visual backbone. For the language backbone, we select models with 7B and 13B parameters for evaluation, respectively. We combine our proposed CIF and PM with the visual backbone and the language backbone into a complete MedVQA model, which generates medical answers based on medical images and question input. Since the image types and QA pairs among the three datasets are relatively similar, we use the same feature extraction strategy. 
\noindent\textbf{Experimental Setup.}
We adopt the open-sourced ViT-B/16 from CLIP~\cite{clip} as our visual backbone and LLaMA~\cite{wu2023pmcllama} as the large language model (LLM) for our MedVQA tasks. Both the visual backbone and the LLM remain frozen during training to leverage their pre-trained capabilities without additional fine-tuning. For the LLM, we select models with 7B and 13B parameters for evaluation. The input image resolution for the visual backbone is set to 224×224 pixels. Since the image types and QA pairs among five datasets are relatively similar, we use the same feature extraction strategy.
To train our causal inference framework (CIF), we employ the AdamW optimizer with a weight decay of 0.05. The initial learning rate is set to 1e-4 , which decays to 1e-7 following a cosine annealing schedule. The model is trained for 100 epochs with a batch size of 16. All models are implemented in PyTorch and trained on 2 NVIDIA GTX 4090 GPUs.

\noindent\textbf{Data Preprocessing.}
To support feature extraction and multimodal integration, we use CLIP’s ViT-B/16 as the visual backbone and LLaMA’s encoder as the language backbone. Both backbones remain frozen during training to preserve their pre-trained knowledge. Visual tokens are generated from input images resized to a resolution of 224×224, where FMM processes them into \( f_{ig} \) and \( f_{il} \) as described above. The \( f_{ig} \) representation provides a holistic understanding of the image, while \( f_{il} \) captures fine-grained, region-specific information, such as anomalies or detailed structures. These embeddings are critical for datasets like VQA-RAD and PathVQA, where precise and diverse visual reasoning is required.

On the textual side, the LLaMA encoder is employed to generate text embeddings from input medical questions. The encoder maps questions into a semantic space that aligns with the extracted visual features, facilitating effective multimodal fusion. These embeddings, together with \( f_{ig} \) and \( f_{il} \), form the input to the causal inference framework, enabling robust reasoning across diverse datasets, including PMC-VQA with its large-scale multimodal data and SLAKE, which requires reasoning across multiple organs and diseases.

\subsection{Main Results}
\label{formats}

\subsubsection{Results on VQA-RAD}
As shown in Table \ref{table:your_table_label}, we achieve the best performance on both open and closed questions by using a language backbone LLaMA with different parameters, 7B and 13B. In particular, we achieve significantly improved performance when using the 13B backbone. 
%For open questions, we improved accuracy by 2.3\%, while for closed questions, we improved accuracy by 1.1\%. Combining the overall VQA-RAD dataset, our method achieved 1.5\% higher than the current SOTA approach.
Specifically, we achieve a 2.3\% accuracy improvement for open questions and a 1.1\% accuracy improvement for closed questions. Combining the overall VQA-RAD dataset, our method outperforms the current state-of-the-art (SOTA) approach by 1.5\%.

\subsubsection{Results on SLAKE}
Compared to all existing MedVQA methods, our method demonstrates the best performance on almost all types of questions. Notably, when using LLaMA-13B as the language backbone, we achieve the highest accuracy across the overall dataset. Our method exhibits a 2.3\% accuracy improvement for open-ended questions compared to the current SOTA. While our accuracy in closed questions is slightly lower than the M2I2 method, our overall dataset performance surpasses the second-place method by 2.3\%.

%%%实验结果+分析+分布+因果作用
%\begin{table}[]
%\centering
%\caption{Results of ablation experiments on the VQA-RAD dataset and SLAKE dataset. ``FDA'' represents the front-door adjustment module.}
%\scalebox{1}
%{
%\begin{tabular}{lcc}
%\toprule 
%\textbf{Method} & \textbf{PMC-VQA} & \textbf{ProbMed} \\
%\midrule
%Chameleon (7B) & 31.0  & 52.8   \\
%LLaVA-Med (v1.5-7B) & 18.9  & 58.5   \\
%MedMax (7B) & 49.0  & 75.8  \\
%\midrule
%Ours (7B) & 49.8  & 70.7  \\
%\bottomrule    
%\end{tabular}}
%\label{bleu}
%\end{table}

\begin{table}[]
\centering
\caption{Distribution and accuracy of closed questions in the SLAKE dataset, where ``Else'' includes those posed with ``Is/Are'', ``Are/Is'', and ``Where''.}
\scalebox{0.85}
{
\begin{tabular}{llllllll}
\toprule
      & \textbf{Does} & \textbf{Is}  & \textbf{Which} & \textbf{Are} & \textbf{Do}  & \textbf{Can} & \textbf{Else} \\
\midrule
\textbf{Train} & 904  & 492 & 242   & 109 & 88  & 67  & 41   \\
\textbf{Val}   & 170  & 115 & 55    & 19  & 29  & 20  & 14   \\
\textbf{Test}  & 172  & 119 & 54    & 21  & 25  & 18  & 7    \\
\midrule
\textbf{Total} & 1246 & 726 & 351   & 149 & 142 & 105 & 62\\
\midrule
\textbf{w/o CIF} & \textbf{87.2} & \textbf{82.4} & \textbf{74.1}   & \textbf{76.2} & \textbf{84.0} & \textbf{77.8} & \textbf{71.4}\\
\textbf{w/ CIF} & \textbf{93.0} & \textbf{89.1} & \textbf{83.3}   & \textbf{85.7} & \textbf{92.0} & \textbf{88.9} & \textbf{85.7}\\
\bottomrule
\label{slakeclosed}
\end{tabular}}
\end{table}
We also conduct statistics on the distribution and accuracy of different types of questions in the SLAKE dataset, and the specific results are shown in Table \ref{slakeclosed} and Table \ref{slakeopen}.
The introduction of the CIF framework brings consistent improvements across all question types in the SLAKE dataset. Specifically, CIF effectively reduces the impact of spurious correlations by aligning visual and textual elements under a causal reasoning framework. This alignment leads to significant accuracy gains, especially for sparse and complex question types such as ``Else'' (+14.3\% for closed questions, +25.0\% for open questions) and ``How'' (+14.3\% for open questions). For common question types like ``Does'' and ``Is'', where the baseline performance is already high, CIF further enhances accuracy by approximately 5-7\%, demonstrating its ability to refine model reasoning even for simpler tasks. These results indicate that CIF not only improves the robustness of the model across diverse question distributions but also enhances its ability to handle complex reasoning and semantic alignment challenges in the MedVQA task.

%, most closed questions can be answered with ``yes'' or ``no'', the accuracy is usually higher, while for those questions that use ``which'', ``what'', or other similar question words, it is often challenging. As shown in Table \ref{slakeopen}, we achieve the highest accuracy for questions posed with ``where''. This shows that the model has excellent judgment ability in location-related questions. 
However, for questions posed in ``How'' (most of which included ``How to'' and ``How many''), the accuracy was relatively low, suggesting that the model still faces challenges in image segmentation and disease inference.

\begin{table}[]
\centering
\caption{Distribution and accuracy of open questions in the SLAKE dataset, where ``Else'' includes those posed with ``In what'' and ``Does''.}
%\small
\setlength{\tabcolsep}{10pt}
\scalebox{0.85}
{
\begin{tabular}{llllll}
\toprule
      & \textbf{What} & \textbf{Which} & \textbf{Where} & \textbf{How} & \textbf{Else} \\
      \midrule
\textbf{Train} & 1640 & 577   & 389   & 342 & 28   \\
\textbf{Val}   & 340  & 127   & 72    & 86  & 6    \\
\textbf{Test}  & 339  & 125   & 89    & 84  & 8    \\
\midrule
\textbf{Total} & 2319 & 829   & 550   & 512 & 42 \\ 
\midrule
\textbf{w/o CIF} & \textbf{81.7} & \textbf{80.8} & \textbf{85.4}   & \textbf{71.4} & \textbf{50.0}\\
\textbf{w/ CIF} & \textbf{91.4} & \textbf{90.4} & \textbf{93.3}   & \textbf{85.7} & \textbf{75.0}\\
\bottomrule
\label{slakeopen}
\end{tabular}}
\end{table}

%%%实验结果+分析+消融所有数据集+加数据集
\begin{table*}[]
\centering
\caption{Results of ablation experiments on five MedVQA datasets. ``CIF'' and ``PM'' represent the causal inference framework and the prompt module, respectively.}
%\setlength{\tabcolsep}{10pt}
%\scalebox{1}
{
\begin{tabular}{lcccccccccccccc}
\toprule \multicolumn{2}{c}{\textbf{Parameters}} & \multicolumn{2}{c}{\textbf{Modules}} &  & \textbf{VQA-RAD} & & & \textbf{SLAKE} & & & \textbf{PathVQA} & &\multirow{2}{*}{\textbf{PMC-VQA}} & \multirow{2}{*}{\textbf{ProbMed}} \\
\cmidrule(lr){1-2} \cmidrule(lr){3-4} \cmidrule(lr){5-7} \cmidrule(lr){8-10} \cmidrule(lr){11-13}
7B & 13B & CIF & PM & Open & Closed & Overall & Open & Closed & Overall & Open & Closed & Overall  \\ \midrule
     \ding{51}  & -  & \ding{55} & \ding{55} & 69.3 & 84.2 & 78.2 & 78.9 & 81.5 & 79.9 & 32.0 & 70.1 & 51.1& 36.0 & 53.0\\
     \ding{51}  &  - & \ding{55} & \ding{51} & 69.3 & 84.6 & 78.5  & 80.3 & 82.7 & 81.2 & 33.6 & 72.3 & 53.0& 38.5& 57.9\\ 
     \ding{51}  &  - & \ding{51} & \ding{55} & 73.7 & 86.4 & 81.4 & 84.8 & 86.5 & 85.5 & 38.3 & 81.7 & 60.1& 43.6 & 66.3\\ 
     \ding{51}  &  - & \ding{51} & \ding{51} & \textbf{74.3} & \textbf{87.1} & \textbf{82.0} & \textbf{90.1} & \textbf{90.4} & \textbf{90.2} & \textbf{40.4} & \textbf{87.9} & \textbf{64.2} & \textbf{44.0} & \textbf{69.4}\\ \midrule
      - & \ding{51} & \ding{55} & \ding{55} & 70.4 & 84.6 & 78.9 & 79.8 & 81.7 & 80.6 & 32.6 & 71.0 & 51.9 &36.6 & 54.0\\
      - & \ding{51} & \ding{55} & \ding{51} & 70.9 & 85.3 & 79.6 & 81.1 & 83.2 & 81.9 & 34.9 & 73.4 & 54.2& 40.8 & 59.1\\ 
      - & \ding{51} & \ding{51} & \ding{55} & 73.7 & 87.1 & 81.8 & 85.9 & 87.5 & 86.5 & 38.9 & 85.6 & 62.3 & 44.2 & 67.2\\ 
      - & \ding{51} & \ding{51} & \ding{51} & \textbf{76.0} & \textbf{87.9} & \textbf{83.1}& \textbf{90.5} & \textbf{91.8} & \textbf{91.0}  & \textbf{41.4} & \textbf{91.5} & \textbf{66.5} & \textbf{46.7} & \textbf{70.7}\\ 
        \bottomrule    
\label{abla}
\end{tabular}}
\end{table*}

\subsubsection{Results on PathVQA}
According to our evaluation results on the PathVQA dataset in Table \ref{table:your_table_label}, our approach outperforms current SOTA by 1.4\% in open questions and 1.2\% in the overall dataset. Notably, the answers to the open-ended questions in the PathVQA dataset are more complex than those in SLAKE and VQA-RAD, which requires models to have a stronger ability to reason and answer creatively, and our model has clear advantages in this respect. Our approach aims to generate answers rather than categorize them from a collection of answers, which enables better performance in terms of open-ended questions.

\subsection{Ablation Study}
\subsubsection{Module analysis} As shown in Table \ref{abla}, we verify the validity of CIF and PM on two language backbones (7B, 13B).

\textbf{Effectiveness of CIF.} When using only CIF, we observe substantial improvements in accuracy across all datasets and question types, particularly for datasets with complex reasoning tasks and high-confounding features such as PathVQA and ProbMed. CIF improves overall accuracy by up to 9.0\% (from 51.1\% to 60.1\%) on PathVQA (7B) and by 13.3\% (from 53.0\% to 66.3\%) on ProbMed (7B). These results indicate that CIF effectively mitigates the impact of spurious correlations, allowing the model to focus on causal relationships between visual inputs, questions, and answers.
CIF also shows consistent improvements in datasets like SLAKE and VQA-RAD, which include a mix of open and closed questions. For example, in SLAKE, CIF improves the overall accuracy from 79.9\% to 85.5\% (7B) and from 80.6\% to 86.5\% (13B), demonstrating its ability to generalize across different question types. The significant improvements for open questions in SLAKE and PathVQA further highlight CIF's capability to address tasks requiring complex reasoning by aligning key visual and textual elements under causal relationships.

\textbf{Effectiveness of PM.}
The prompt module (PM) also leads to noticeable improvements, particularly in datasets where the complexity of medical questions and textual context poses challenges for large language models (LLMs). PM enhances the model’s understanding of complex medical questions by providing structured guidance through prompts, which dynamically adapt to the context of each question. For example, on ProbMed, PM increases overall accuracy from 53.0\% to 57.9\% (7B) and from 54.0\% to 59.1\% (13B). Similarly, in PathVQA, PM improves the overall accuracy by 1.9\% (from 51.1\% to 53.0\%) for 7B and by 2.3\% (from 51.9\% to 54.2\%) for 13B.
The impact of PM is particularly pronounced in datasets with diverse question types, such as SLAKE and PMC-VQA, where it facilitates better alignment between textual inputs and the causal features extracted by the model. For example, in SLAKE, PM increases overall accuracy from 79.9\% to 81.2\% (7B) and from 80.6\% to 81.9\% (13B), demonstrating its effectiveness in improving the model’s ability to interpret nuanced medical concepts. 
On PMC-VQA, PM increases accuracy from 36.0\% to 38.5\% (7B) and from 36.6\% to 40.8\% (13B).
On VQA-RAD, PM provides a modest 0.3\% overall accuracy gain (from 78.2\% to 78.5\% for 7B). This limited impact is due to VQA-RAD's relatively structured question-answer format, where most questions adhere to consistent patterns, reducing the need for additional prompt-based guidance.

\textbf{Combined Effect of CIF and PM.}
The best performance is achieved when CIF and PM are used together, resulting in the highest accuracy across all datasets.
On ProbMed, the combined approach achieves an overall accuracy of 69.4\% (7B) and 70.7\% (13B), representing improvements of 16.4\% and 16.7\%, respectively, compared to the baseline without CIF and PM.
On PathVQA, accuracy increases to 64.2\% (7B) and 66.5\% (13B), highlighting the importance of combining causal alignment (CIF) with the semantic guidance provided by PM.
On PMC-VQA, while the 0.4\% improvement for 7B may seem modest, it remains meaningful given the dataset’s large scale and diversity. With 227K QA pairs, even a small percentage gain translates to tangible improvements across hundreds of cases.
This complementary effect can be attributed to the distinct roles of CIF and PM in the MedVQA pipeline:
CIF eliminates confounders by aligning visual and textual features under causal relationships, reducing the model's reliance on spurious correlations. This ensures that the features extracted by the model are relevant and task-specific.
PM further enhances the utilization of these causal features by providing question-specific guidance, allowing the LLMs to focus on contextually important information. While CIF ensures that the extracted features are causally valid, PM makes these features more accessible and interpretable for the LLMs, particularly in datasets with complex or diverse medical questions.

%When using only CIF, we achieve comparability improvements in the overall dataset(increased by 3.2\%, 1.9\%), indicating that CIF eliminates the effects of confounders. Given that each image corresponds to distinct QA pairs, and confounders are determined based on different questions, this highlights CIF's ability to mitigate the effect of the relativity of confounders. When using only PM, there is a certain increase in accuracy (increased by 0.3\%, 0.7\%), indicating that PM provides more instructive information for LLMs. 
%This indicates that PM improves the model's perception of complex inputs and medical concepts. The best performance is achieved when CIF and PM are used together (increased by 3.8\%, 4.2\%). We use PM as auxiliary information to alleviate the need for alignment between the two modals, enabling LLMs to utilize causal features more effectively. CIF can eliminate confounders, but the complexity of LLMs may lead to difficulties in understanding the specific context and questions, so causal features may not be fully utilized. Fortunately, PM provides more targeted information, making the model more flexible to understand and answer questions. Therefore, the better performance of the combination of CIF and PM is due to their complementary roles.

%%%实验结果+分析，回应主干+效果+加数据集
\begin{table}[]
\centering
\caption{Performance comparison of different vision and text encoders.}
\scalebox{1}{
\begin{tabular}{llccc}
\toprule
\textbf{LB} & \textbf{VB} & \textbf{CIF+PM} & \textbf{PMC-VQA} & \textbf{ProbMed}  \\
\midrule
\multirow{6}{*}{BERT}    &         ResNet-101       & \ding{55}& 16.3  & 31.2   \\
 &         ResNet-101       & \ding{51}& 20.4  & 37.8   \\
 &         MEVF    & \ding{55}  & 20.9  & 35.5   \\
 &         MEVF       & \ding{51}& \underline{26.6}  & 42.4   \\
   & CLIP ViT-B/16         & \ding{55}& 21.7  & \underline{43.7}   \\
                              & CLIP ViT-B/16         & \ding{51}& \textbf{30.4}  & \textbf{51.0}   \\
\midrule
\multirow{6}{*}{DeBERTa}     &         ResNet-101       &\ding{55} & 25.1  & 38.9   \\
 &         ResNet-101       & \ding{51}& 27.4  & 42.1   \\
 &         MEVF       & \ding{55}& 30.5  & 40.3   \\
 &         MEVF       & \ding{51}& \underline{34.7}  & 45.6   \\
   & CLIP ViT-B/16         & \ding{55}& 33.7  & \underline{50.4}   \\
                              & CLIP ViT-B/16         & \ding{51}& \textbf{39.2}  & \textbf{57.2}   \\
\midrule
\multirow{6}{*}{LLaMA-7B}      &         ResNet-101       & \ding{55}& 23.4  & 41.7   \\
 &         ResNet-101       & \ding{51}& 28.5  & 50.8   \\
 &         MEVF       & \ding{55}& 30.2  & 44.1   \\
 &         MEVF       & \ding{51}& 35.6  & \underline{55.2}   \\
   & CLIP ViT-B/16         & \ding{55}& \underline{36.0}  & 53.0   \\
                              & CLIP ViT-B/16         & \ding{51}& \textbf{44.0}  & \textbf{69.4}   \\
\bottomrule
\end{tabular}}
\label{tab:comparison}
\end{table}

\subsubsection{Impact Analysis of Encoders}
Table \ref{tab:comparison} presents the performance comparison of different language backbones (LB), vision backbones (VB), and their combinations with the proposed CIF+PM framework on the PMC-VQA and ProbMed datasets. The results highlight the significant role of CIF+PM in improving the performance across different backbones.

\textbf{Comparison Across Vision Backbones.}
Different vision backbones show varying levels of performance, with CLIP ViT-B/16 consistently outperforming ResNet-101 and MEVF. For instance, in the BERT setting without CIF+PM, CLIP achieves 21.7\% and 43.7\% on PMC-VQA and ProbMed, which is higher than ResNet-101 (16.3\% and 31.2\%) and MEVF (20.9\% and 35.5\%). This demonstrates that CLIP provides a stronger visual representation, which becomes even more effective when combined with CIF+PM. In the DeBERTa and LLaMA-7B settings, similar trends are observed, with CLIP + CIF+PM achieving the best results across both datasets, such as 39.2\% on PMC-VQA and 57.2\% on ProbMed for the DeBERTa, and 44.0\% and 69.4\% for LLaMA-7B.

\textbf{Comparison Across Language Backbones.}
Language backbones also significantly impact performance. Among the three LBs, LLaMA-7B consistently achieves the best results, especially when combined with CIF+PM. For example, in the CLIP setting with CIF+PM, LLaMA-7B achieves the highest scores of 44.0\% on PMC-VQA and 69.4\% on ProbMed, outperforming DeBERTa (39.2\% and 57.2\%) and BERT (30.4\% and 51.0\%). These results suggest that LLaMA(7B), as a larger and more powerful language model, can better leverage the causal features extracted by CIF and the structured guidance provided by PM, particularly for complex datasets like ProbMed.

\textbf{Effectiveness of CIF+PM.}
Across all combinations of LB and VB, the inclusion of CIF+PM consistently leads to substantial performance gains. For example, with the LLaMA-7B + CLIP backbone, CIF+PM improves the accuracy on PMC-VQA from 36.0\% to 44.0\% and on ProbMed from 53.0\% to 69.4\%. 
The results further highlight the complementary roles of CIF and PM. Without CIF+PM, even the best-performing backbones (e.g., LLaMA-7B + CLIP) achieve only 36.0\% on PMC-VQA and 53.0\% on ProbMed. However, the addition of CIF+PM leads to significant accuracy improvements, with increases of 8.0\% on PMC-VQA and 16.4\% on ProbMed. This underscores the effectiveness of CIF in mitigating the impact of confounders and aligning visual and textual elements under causal reasoning. Simultaneously, PM enhances the utilization of these causal features by providing structured, question-specific guidance, enabling the language model to focus on relevant context and generate accurate answers.

%\textbf{Dataset-Specific Observations.}
%PMC-VQA: The results on PMC-VQA demonstrate that CIF+PM improves both visual-textual alignment and reasoning capabilities, particularly in tasks requiring precise understanding of medical images. For example, the combination of CIF+PM with MEVF and CLIP consistently improves accuracy across all language backbones.
%ProbMed: The improvements on ProbMed are even more pronounced, reflecting the effectiveness of CIF+PM in handling datasets with high-confounding features and complex medical reasoning tasks. For example, the accuracy improvement from 53.0\% to 69.4\% with LLaMA-7B + CLIP highlights the ability of our approach to generalize to challenging datasets.

\subsubsection{Parameter analysis} In MedVQA, we adopt the causal inference method to construct two modal mediators $\mathcal{M}_i$ and $\mathcal{M}_q$. To gain insight into the reasons for the performance improvement, we introduce an experiment that bypasses the Front-Door Adjustment (FDA) module by experimenting with extracted $\mathcal{M}_i$ and $\mathcal{M}_q$ directly. This is designed to investigate whether the performance improvement is due to confounding removal or the number of parameters introduced. Results in Table \ref{abla2} show that for open-ended questions in the VQA-RAD dataset, using the FDA module improves accuracy by 3.4\% compared to not using it. For closed questions, the increase is 1.4\% and the overall performance improves by 2.2\%. In the SLAKE dataset, the performances improve by 3.1\%, 2.4\%, and 2.8\%, respectively.
However, as shown in Table \ref{abla}, the performance in VQA-RAD without CIF is 69.3\%, 84.6\%, and 78.5\%. After using CIF, even without using the FDA module, there is a slight performance improvement of 1.6\%, 1.1\%, and 1.3\%, due to the role of the mediators. As shown in Fig. \ref{fig:ind}, although the process of building mediators is not a complete causal inference process, it is essentially a feature fusion mechanism. This means that $\mathcal{M}_i$ and $\mathcal{M}_q$ are more information-rich than the original features, resulting in a slight performance improvement. However, with the use of FDA, the performance improvements reached 4.0\%, 2.5\%, and 3.5\%, respectively, far exceeding the effects of only using mediators. This clearly shows that the significant improvement in performance is almost entirely due to the causal inference process and not just the introduction of the number of parameters. At the same time, this highlights the need to ensure the integrity of the CIF module.

\begin{table}[]
\centering
\caption{Results of ablation experiments on the VQA-RAD dataset and SLAKE dataset. ``FDA'' represents the front-door adjustment module.}
\scalebox{0.8}
%\scalebox{1}
{
\begin{tabular}{ccccccc}
\toprule \multirow{2}{*}{\textbf{FDA}} & &\textbf{VQA-RAD}& & &\textbf{SLAKE}& \\
\cmidrule(lr){2-4} \cmidrule(lr){5-7}
~ & Open & Closed & Overall & Open & Closed & Overall \\ \midrule
      \ding{55} & 70.9  & 85.7 & 79.8 & 87.0 & 88.0 & 87.4 \\
     \ding{51}  & \textbf{74.3}  & \textbf{87.1} & \textbf{82.0} & \textbf{90.1} & \textbf{90.4} & \textbf{90.2} \\  
        \bottomrule    
\label{abla2}
\end{tabular}}
\end{table}

%%%实验结果+分析+回应指标+增数据集

\begin{table}[]
\centering
\caption{Metric analysis experiments on the SLAKE dataset and PathVQA dataset.}
\scalebox{0.85}{
\begin{tabular}{lcccc}
\toprule 
\multirow{2}{*}{\textbf{Method}} & \multicolumn{2}{c}{\textbf{SLAKE}} & \multicolumn{2}{c}{\textbf{PathVQA}} \\
\cmidrule(lr){2-3} \cmidrule(lr){4-5}
~ & \textbf{BLEU-1} & \textbf{F1} & \textbf{BLEU-1} & \textbf{F1}  \\ 
\midrule
Med-Flamingo\cite{flamingo} & 21.51  & 23.66  & 33.38 & 34.01 \\
RadFM\cite{radfm}         & 81.66  & 82.38 & 24.83 & 25.20 \\
LLaVA-Med\cite{llava-med} & 76.95  & 77.30  & 46.42 & 47.08 \\
Uni-Med\cite{unimed} & 82.12  & 83.07  & 58.07 & 58.74 \\
\textbf{Ours} & \textbf{85.40}  & \textbf{85.25}  & \textbf{63.93} & \textbf{62.72} \\
\bottomrule    
\end{tabular}}
\label{bleu}
\end{table}

\begin{figure}[]
 \centering
\includegraphics[width=1\linewidth]{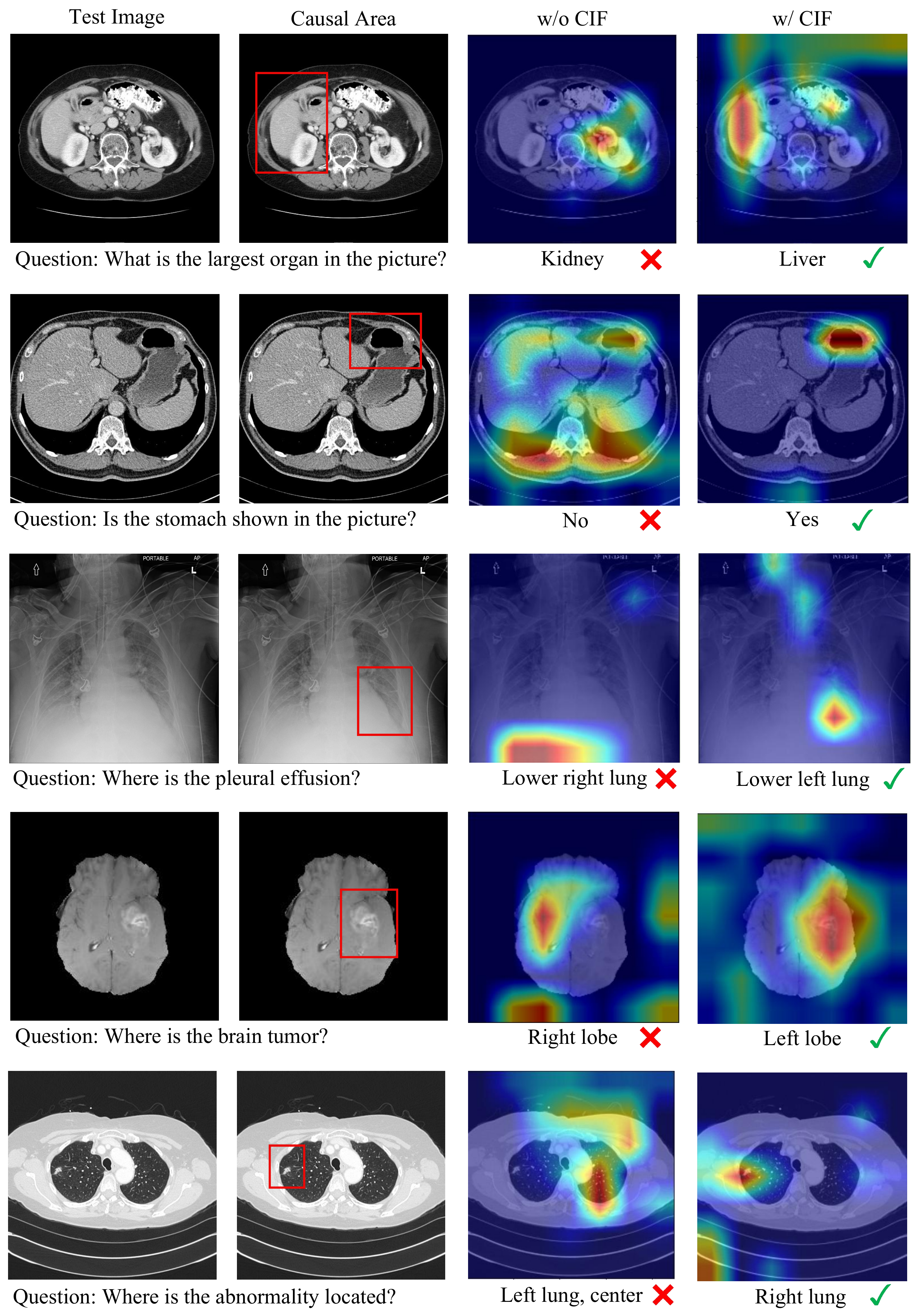}
 \caption{Visualization Results of CIF on Lung (X-ray, CT), Brain (MRI), and Abdomen (CT) Imaging.
 }
 \label{fig:grad}
\end{figure}

\begin{figure*}[!ht]
 \centering
\includegraphics[width=1\linewidth]{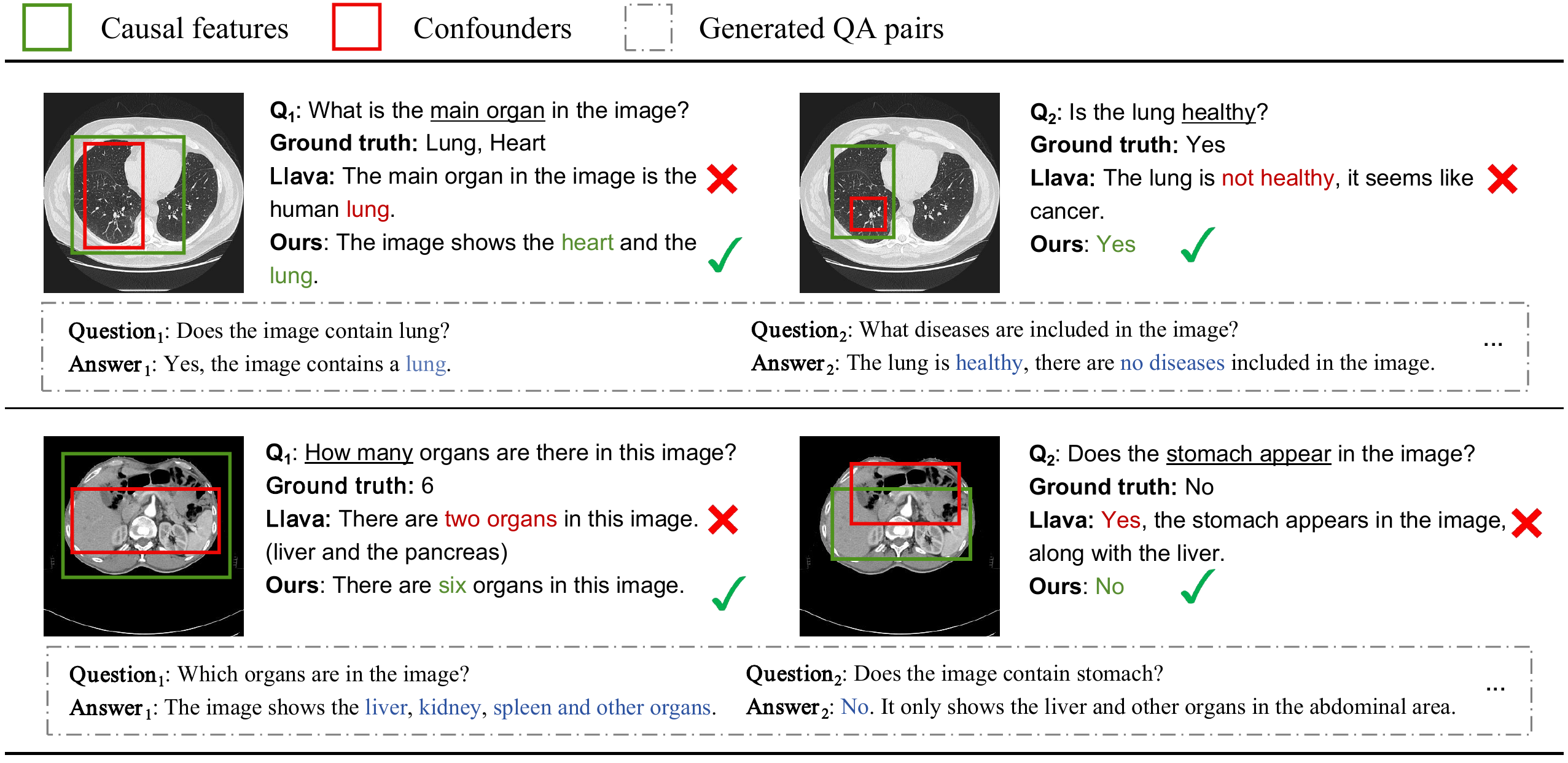}
 \caption{Qualitative results on SLAKE. Combined with the causal inference and prompt modules, our approach performs well in MedVQA. 
 The blue font represents the part of the prompt text that guides the correct answer.
 %Bold green indicates text that directly guides the model to answer the question. 
 %See more examples in \textbf{Appendix}.
 }
 \label{fig:result}
\end{figure*}

\subsection{Metric Analysis}
Table \ref{bleu} shows the performance comparison of different methods on the SLAKE and PathVQA datasets using BLEU-1 and F1 metrics. Our method achieves the highest scores across both datasets, with BLEU-1 and F1 scores of 91.40 and 85.25 on SLAKE, and 66.93 and 62.72 on PathVQA, respectively. Compared to the strongest baseline Uni-Med, our method improves BLEU-1 by 3.28 and 5.86 points and F1 by 2.18 and 3.98 points on SLAKE and PathVQA, respectively. These results demonstrate the superior ability of our approach to generate precise and semantically aligned answers. The causal inference framework (CIF) effectively mitigates confounding effects and aligns visual and textual features under causal relationships, while the prompt module (PM) enhances the model's understanding of complex medical questions by providing structured guidance. 

\subsection{Qualitative Analysis}
Fig. \ref{fig:grad} shows the significant changes in the model's attention distribution and corresponding answer accuracy before and after applying the CIF module across various medical imaging modalities, including lung (X-ray/CT), brain (MRI), and abdomen (CT) images. On lung images, when answering the question ``Where is the pleural effusion?'', the model without CIF incorrectly focused on the lower right lung instead of the true lesion area in the lower left lung. This error is caused by confounding effects, which misled the model to associate frequent statistical correlations with the answer. After introducing CIF, the model correctly attended to the pleural effusion region in the lower left lung and provided the correct answer, demonstrating CIF's ability to eliminate spurious correlations and guide the model to identify causally relevant regions based on causal reasoning. On brain MRI images, CIF effectively mitigates the influence of confounding regions in complex brain imaging tasks, significantly improving the model's focus on causally relevant areas. Similarly, on abdominal CT images, when asked "What is the largest organ in the picture?", the model without CIF mistakenly focused on the irrelevant organ and produced an incorrect answer. In contrast, the model with CIF accurately generated the correct answer, demonstrating CIF’s capability to separate causally relevant features from irrelevant ones, even in complex multi-organ imaging scenarios. 

Overall, without CIF, the model is often misled by confounding effects, dispersing its attention to non-causal regions and producing incorrect answers. With CIF, the model's attention is effectively concentrated on regions directly relevant to the question, resulting in significantly improved answer accuracy. These results validate CIF's critical role in mitigating the influence of spurious correlations, enabling the model to focus on causally relevant regions, and strengthening the causal alignment between visual and textual modalities. 
%CIF not only reduces the model's over-reliance on statistical correlations but also consistently demonstrates robust performance across diverse medical imaging modalities, including pulmonary X-ray, brain MRI, and abdominal CT. These qualitative results highlight CIF's significant advantages in eliminating confounding effects, enhancing visual-text alignment, and improving answer accuracy, thereby providing strong support for the model's interpretability and robustness.

Fig.~\ref{fig:result} shows the qualitative analysis of our proposed method. We select two images, each corresponding to two questions. In both cases, there is a situation where the region of confounder that leads the model to make an incorrect prediction, is actually the real cause of another question. In the first image, Llava~\cite{liu2023llava} ignores the presence of the heart and misdiagnoses the healthy image as cancer. In the second image, Llava has trouble recognizing organs. For different questions with each image, CIF captures different confounding factors. By eliminating spurious associations caused by relative confounders, CIF guides the model to provide accurate answers. In addition, some of the QA pairs generated by PM include the correct answers. By controlling the form of the generated QA pairs, PM is broadly applicable and provides useful information for both open and closed questions. These examples highlight the critical role of causal inference and prompt strategy in MedVQA.

\section{Discussion}
\subsection{Causal inference}
In our work, we leveraged the front-door adjustment (FDA) method from causal inference to address confounding effects in the MedVQA task. As shown in Fig.~\ref{fig:cau}, the FDA approach introduces two mediating variables, $\mathcal{M}_i$ and $\mathcal{M}_q$, which effectively cut off the backdoor paths. This adjustment aims to eliminate the influence of confounders and allows for a more accurate estimation of the causal effect.We implemented the FDA in MedVQA, by intervening in the variables \( I \) and \( Q \) to block the backdoor paths. This approach is well-established in causal inference and has demonstrated success in various applications. To solidify our research's theoretical foundation, we simplified the probability formula in Eq.\ref{eq:fd44} through mathematical analysis.

Our investigation delved into the mathematical principles underlying causal inference, followed by extensive experiments and analyses. The critical aspect of applying causal inference lies in realizing the probability formulas derived from theoretical foundations, bridging theory and practice. This process necessitates translating theoretical formulas into practical sub-networks.
After numerous attempts, we constructed mediating variables based on the attention mechanism (as detailed in Fig \ref{fig:ind}). 
By utilizing the FDA formula, we decomposed the causal effect into the product of the distribution of the mediating variable and the distribution of the target variable, enabling a more precise estimation of the causal effect. This method underscores the potential of causal inference techniques in enhancing the reliability and accuracy of outcomes in MedVQA tasks.

\subsection{Prompt module}
In our approach, we employed prompts as auxiliary information to address the alignment challenges between different modalities. This integration allows large language models (LLMs) to leverage causal features more effectively. While causal inference frameworks (CIF) can eliminate confounders, the inherent complexity of LLMs can sometimes hinder their ability to fully grasp the specific context and questions. As a result, the causal features may not be optimally utilized.

The introduction of prompt modules (PM) offers a solution by providing more targeted information, enhancing the model's flexibility in understanding and responding to questions. The prompts serve as a bridge, facilitating better comprehension and alignment within the LLMs. We observed that the combination of CIF and PM yielded superior performance, which can be attributed to their complementary roles. CIF mitigates the impact of confounders, ensuring a clearer causal pathway, while PM supplies the necessary contextual information, optimizing the LLM's interpretative and response capabilities.
In conclusion, this synergy between causal inference and prompt modules underscores a significant advancement in MedVQA tasks. 
%By integrating these techniques, we can achieve a more robust and accurate model, capable of effectively handling the complexities of medical visual question answering. This combined approach not only enhances the reliability of the causal inferences drawn but also ensures that the LLMs can more precisely and flexibly navigate the nuanced information presented in medical contexts.

\subsection{Limitation}
Despite the promising results achieved with our current approach, there are notable limitations that warrant discussion. Our method of constructing mediating variables, although effective, may still be limited in its ability to fully eliminate confounding effects. The complexity and variability inherent in medical visual question answering tasks mean that there may be residual confounders that our current model does not completely account for. This limitation underscores the need for continuous refinement and improvement of our mediating variable construction techniques.
Furthermore, while our approach has successfully leveraged causal inference principles to enhance model performance, there remains an ongoing challenge in quantifying the impact of eliminating confounders. Future work will focus on calculating the causal effect to more precisely measure the extent to which confounding influences have been mitigated. 

%In summary, while our integration of front-door adjustment and prompt modules has demonstrated significant potential, it is clear that further advancements are needed to fully realize the benefits of causal inference in MedVQA. By addressing these limitations and continuing to refine our methods, we aim to develop more sophisticated models that can more effectively handle the complexities of medical visual question answering, ultimately leading to more accurate and reliable outcomes.

\section{Conclusion}
In this work, we propose a novel causal inference framework to learn the causal structure of multi-modal medical information from the perspective of causal representation. We emphasize the correlation of confounders, reduce their impact by employing multi-variate resampling front-door adjustments, and finally achieve true causal associations between MedVQA data.
In addition, we design a prompt module to help the LLMs understand the context of the questions and the visual scene. It allows the model to better generalize to multi-modal medical data while reducing the deployment cost of the model. By combining the two with pre-trained visual and language backbones, we have significantly improved the accuracy of MedVQA, which is expected to promote deep learning research and the application of multi-modal medical information.

\bibliographystyle{IEEEtran}
	\bibliography{bare_adv}

@String(CVPR  = {IEEE Conf. Comput. Vis. Pattern Recog.})

@String(CVPR  = {CVPR})

@inproceedings{guo2023from,
  title={From images to textual prompts: Zero-shot visual question answering with frozen large language models},
  author={Jiaxian Guo et al.},
  booktitle={Proceedings of the IEEE/CVF conference on computer vision and pattern recognition},
  pages={10867--10877},
  year={2023}
}

@article{huang2023dual,
  title={A dual-attention learning network with word and sentence embedding for medical visual question answering},
  author={Huang, Xiaofei and Gong, Hongfang},
  journal={IEEE Transactions on Medical Imaging},
  year={2023},
  publisher={IEEE}
}

@article{tiong2022plug,
  title={Plug-and-Play VQA: Zero-shot VQA by Conjoining Large Pretrained Models with Zero Training},
  author={Tiong, Anthony Meng Huat and Li, Junnan and Li, Boyang and Savarese, Silvio and Hoi, Steven CH},
  journal={arXiv preprint arXiv:2210.08773},
  year={2022}
}

@ARTICLE{yx1,
  author={Yang, Xun and Wang, Shanshan and Dong, Jian and Dong, Jianfeng and Wang, Meng and Chua, Tat-Seng},
  journal={IEEE Transactions on Image Processing}, 
  title={Video Moment Retrieval With Cross-Modal Neural Architecture Search}, 
  year={2022},
  volume={31},
  number={},
  pages={1204-1216},
  doi={10.1109/TIP.2022.3140611}}

@inproceedings{yx2,
author = {Yang, Xun and Feng, Fuli and Ji, Wei and Wang, Meng and Chua, Tat-Seng},
title = {Deconfounded Video Moment Retrieval with Causal Intervention},
year = {2021},
isbn = {9781450380379},
publisher = {Association for Computing Machinery},
address = {New York, NY, USA},
doi = {10.1145/3404835.3462823},
booktitle = {Proceedings of the 44th International ACM SIGIR Conference on Research and Development in Information Retrieval},
pages = {1–10},
numpages = {10},
location = {Virtual Event, Canada},
series = {SIGIR '21}
}

@article{yx3,
author = {Li, Kun and Li, Jiaxiu and Guo, Dan and Yang, Xun and Wang, Meng},
title = {Transformer-Based Visual Grounding with Cross-Modality Interaction},
year = {2023},
issue_date = {November 2023},
publisher = {Association for Computing Machinery},
address = {New York, NY, USA},
volume = {19},
number = {6},
issn = {1551-6857},
doi = {10.1145/3587251},
journal = {ACM Trans. Multimedia Comput. Commun. Appl.},
month = may,
articleno = {183},
numpages = {19}
}

@InProceedings{clip,
  title = 	 {Learning Transferable Visual Models From Natural Language Supervision},
  author =       {Radford Alec  et al.},
  booktitle = 	 {Proceedings of the 38th International Conference on Machine Learning},
  pages = 	 {8748--8763},
  year = 	 {2021},
  editor = 	 {Meila, Marina and Zhang, Tong},
  volume = 	 {139},
  series = 	 {Proceedings of Machine Learning Research},
  month = 	 {18--24 Jul},
  publisher =    {PMLR}
}

@article{probme,
  title={Worse than Random? An Embarrassingly Simple Probing Evaluation of Large Multimodal Models in Medical VQA},
  author={Yan, Qianqi and He, Xuehai and Yue, Xiang and Wang, Xin Eric},
  journal={arXiv preprint arXiv:2405.20421},
  year={2024}
}

@article{pmcvqa,
  title={Pmc-vqa: Visual instruction tuning for medical visual question answering},
  author={Zhang, Xiaoman and Wu, Chaoyi and Zhao, Ziheng and Lin, Weixiong and Zhang, Ya and Wang, Yanfeng and Xie, Weidi},
  journal={arXiv preprint arXiv:2305.10415},
  year={2023}
}

@article{zhang2021gcastle,
  title={gcastle: A python toolbox for causal discovery},
  author={Keli Zhang et al.},
  journal={arXiv preprint arXiv:2111.15155},
  year={2021}
}

@inproceedings{chen2022m3ae,
  title={Multi-modal masked autoencoders for medical vision-and-language pre-training},
  author={Chen Zhihong et al.},
  booktitle={International Conference on Medical Image Computing and Computer-Assisted Intervention},
  pages={679--689},
  year={2022},
  organization={Springer}
}

@article{zhang2023pmcvqa,
      title={PMC-VQA: Visual Instruction Tuning for Medical Visual Question Answering}, 
      author={Xiaoman Zhang et al.},
      year={2023},
      journal={arXiv preprint arXiv:2305.10415},
}

@inproceedings{lin2023pmc,
  title={Pmc-clip: Contrastive language-image pre-training using biomedical documents},
  author={Lin Weixiong et al.},
  booktitle={International Conference on Medical Image Computing and Computer-Assisted Intervention},
  pages={525--536},
  year={2023},
  organization={Springer}
}

@article{yuan2022auto,
  title={Auto IV: Counterfactual Prediction via Automatic Instrumental Variable Decomposition},
  author={Yuan, Junkun et al.},
  journal={ACM Transactions on Knowledge Discovery from Data (TKDD)},
  volume={16},
  number={4},
  pages={1--20},
  year={2022},
  publisher={ACM New York, NY}
}

@article{wu2023pmcllama,
  title={PMC-LLaMA: toward building open-source language models for medicine},
  author={Wu, Chaoyi and Lin, Weixiong and Zhang, Xiaoman and Zhang, Ya and Xie, Weidi and Wang, Yanfeng},
  journal={Journal of the American Medical Informatics Association},
  pages={ocae045},
  year={2024},
  publisher={Oxford University Press}
}

@article{Zang2023DiscoveringTR,
  title={Discovering the Real Association: Multimodal Causal Reasoning in Video Question Answering},
  author={Chuanqi Zang and Hanqing Wang and Mingtao Pei and Wei Liang},
  journal={2023 IEEE/CVF Conference on Computer Vision and Pattern Recognition (CVPR)},
  year={2023},
  pages={19027-19036}
}

@article{Lu2023MultiscaleFE,
  title={Multiscale Feature Extraction and Fusion of Image and Text in VQA},
  author={Siyu Lu and Yue Ding and Mingzhe Liu and Zhengtong Yin and Lirong Yin and Wenfeng Zheng},
  journal={International Journal of Computational Intelligence Systems},
  year={2023},
  volume={16}
}

@inproceedings{flamingo,
  title={Med-flamingo: a multimodal medical few-shot learner},
  author={Moor, Michael and Huang, Qian and Wu, Shirley and Yasunaga, Michihiro and Dalmia, Yash and Leskovec, Jure and Zakka, Cyril and Reis, Eduardo Pontes and Rajpurkar, Pranav},
  booktitle={Machine Learning for Health (ML4H)},
  pages={353--367},
  year={2023},
  organization={PMLR}
}

@article{radfm,
  title={Towards generalist foundation model for radiology},
  author={Wu, Chaoyi and Zhang, Xiaoman and Zhang, Ya and Wang, Yanfeng and Xie, Weidi},
  journal={arXiv preprint arXiv:2308.02463},
  year={2023}
}

@article{Pearl2009CausalII,
  title={Causal inference in statistics: An overview},
  author={Judea Pearl},
  journal={Statistics Surveys},
  year={2009},
  volume={3},
  pages={96-146}
}

@article{llava-med,
  title={Llava-med: Training a large language-and-vision assistant for biomedicine in one day},
  author={Li, Chunyuan and Wong, Cliff and Zhang, Sheng and Usuyama, Naoto and Liu, Haotian and Yang, Jianwei and Naumann, Tristan and Poon, Hoifung and Gao, Jianfeng},
  journal={Advances in Neural Information Processing Systems},
  volume={36},
  year={2024}
}

@inproceedings{liu2023llava,
 author = {Liu, Haotian and Li, Chunyuan and Wu, Qingyang and Lee, Yong Jae},
 booktitle = {Advances in Neural Information Processing Systems},
 pages = {34892--34916},
 publisher = {Curran Associates, Inc.},
 title = {Visual Instruction Tuning},
 volume = {36},
 year = {2023}
}

@article{unimed,
  title={Uni-Med: A Unified Medical Generalist Foundation Model For Multi-Task Learning Via Connector-MoE},
  author={Zhu, Xun and Hu, Ying and Mo, Fanbin and Li, Miao and Wu, Ji},
  journal={arXiv preprint arXiv:2409.17508},
  year={2024}
}

@inproceedings{zhan2023miccai,
  title={Debiasing medical visual question answering via counterfactual training},
  author={Zhan, Chenlu and Peng, Peng and Zhang, Hanrong and Sun, Haiyue and Shang, Chunnan and Chen, Tao and Wang, Hongsen and Wang, Gaoang and Wang, Hongwei},
  booktitle={International Conference on Medical Image Computing and Computer-Assisted Intervention},
  pages={382--393},
  year={2023},
  organization={Springer}
}

@inproceedings{ye2024causal,
  title={A Causal Approach to Mitigate Modality Preference Bias in Medical Visual Question Answering},
  author={Ye, Shuchang and Naseem, Usman and Meng, Mingyuan and Feng, Dagan and Kim, Jinman},
  booktitle={Proceedings of the First International Workshop on Vision-Language Models for Biomedical Applications},
  pages={13--17},
  year={2024}
}

@article{cai2024counterfactual,
  title={Counterfactual causal-effect intervention for interpretable medical visual question answering},
  author={Cai, Linqin and Fang, Haodu and Xu, Nuoying and Ren, Bo},
  journal={IEEE Transactions on Medical Imaging},
  year={2024},
  publisher={IEEE}
}

@inproceedings{40cb06d16fd1450ea39bfd13d43e9c9f,
  title={Slake: A semantically-labeled knowledge-enhanced dataset for medical visual question answering},
  author={Liu, Bo and Zhan, Li-Ming and Xu, Li and Ma, Lin and Yang, Yan and Wu, Xiao-Ming},
  booktitle={2021 IEEE 18th International Symposium on Biomedical Imaging (ISBI)},
  pages={1650--1654},
  year={2021},
  organization={IEEE}
}

@article{He2020PathVQA3Q,
  title={PathVQA: 30000+ Questions for Medical Visual Question Answering},
  author={Xuehai He and Yichen Zhang and Luntian Mou and Eric P. Xing and Pengtao Xie},
  journal={ArXiv},
  year={2020},
  volume={abs/2003.10286}
}

@article{Lau2018DescriptorA,
  title={A dataset of clinically generated visual questions and answers about radiology images},
  author={Lau, Jason J and Gayen, Soumya and Ben Abacha, Asma and Demner-Fushman, Dina},
  journal={Scientific data},
  volume={5},
  number={1},
  pages={1--10},
  year={2018},
  publisher={Nature Publishing Group}
}

@inproceedings{Peng2018UMassAI,
  title={UMass at ImageCLEF Medical Visual Question Answering (Med-VQA) 2018 Task.},
  author={Peng, Yalei and Liu, Feifan and Rosen, Max P},
  booktitle={CLEF (working notes)},
  pages={1--9},
  year={2018}
}

@inproceedings{Zhou2018EmployingIA,
  title={Employing Inception-Resnet-v2 and Bi-LSTM for Medical Domain Visual Question Answering.},
  author={Zhou, Yangyang and Kang, Xin and Ren, Fuji},
  booktitle={CLEF (working notes)},
  pages={1--11},
  year={2018}
}

@inproceedings{Abacha2018NLMAI,
  title={NLM at ImageCLEF 2018 Visual Question Answering in the Medical Domain.},
  author={Abacha, Asma Ben and Gayen, Soumya and Lau, Jason J and Rajaraman, Sivaramakrishnan and Demner-Fushman, Dina},
  booktitle={CLEF (working notes)},
  pages={1--10},
  year={2018}
}

@article{Simonyan2014VeryDC,
  title={Very Deep Convolutional Networks for Large-Scale Image Recognition},
  author={Karen Simonyan and Andrew Zisserman},
  journal={CoRR},
  year={2014},
  volume={abs/1409.1556}
}

@article{He2015DeepRL,
  title={Deep Residual Learning for Image Recognition},
  author={Kaiming He and X. Zhang and Shaoqing Ren and Jian Sun},
  journal={2016 IEEE Conference on Computer Vision and Pattern Recognition (CVPR)},
  year={2015},
  pages={770-778}
}

@inproceedings{Nguyen2019OvercomingDL,
  title={Overcoming data limitation in medical visual question answering},
  author={Nguyen, Binh D and Do, Thanh-Toan and Nguyen, Binh X and Do, Tuong and Tjiputra, Erman and Tran, Quang D},
  booktitle={Medical Image Computing and Computer Assisted Intervention--MICCAI 2019},
  pages={522--530},
  year={2019},
  organization={Springer}
}

@inproceedings{Masci2011StackedCA,
  title={Stacked convolutional auto-encoders for hierarchical feature extraction},
  author={Masci, Jonathan and Meier, Ueli and Cire{\c{s}}an, Dan and Schmidhuber, J{\"u}rgen},
  booktitle={Artificial Neural Networks and Machine Learning--ICANN 2011},
  pages={52--59},
  year={2011},
  organization={Springer}
}

@article{Vuorio2019MultimodalMM,
  title={Multimodal model-agnostic meta-learning via task-aware modulation},
  author={Vuorio, Risto and Sun, Shao-Hua and Hu, Hexiang and Lim, Joseph J},
  journal={Advances in neural information processing systems},
  volume={32},
  year={2019}
}

@inproceedings{Zhan2020MedicalVQ,
  title={Medical visual question answering via conditional reasoning},
  author={Zhan, Li-Ming and Liu, Bo and Fan, Lu and Chen, Jiaxin and Wu, Xiao-Ming},
  booktitle={Proceedings of the 28th ACM International Conference on Multimedia},
  pages={2345--2354},
  year={2020}
}

@inproceedings{Liu2021ContrastivePA,
  title={Contrastive pre-training and representation distillation for medical visual question answering based on radiology images},
  author={Liu, Bo and Zhan, Li-Ming and Wu, Xiao-Ming},
  booktitle={Medical Image Computing and Computer Assisted Intervention--MICCAI 2021},
  pages={210--220},
  year={2021},
  organization={Springer}
}

@inproceedings{Bareinboim2011ControllingSB,
  title={Controlling selection bias in causal inference},
  author={Bareinboim, Elias and Pearl, Judea},
  booktitle={Artificial Intelligence and Statistics},
  pages={100--108},
  year={2012},
  organization={PMLR}
}

@article{Wang2020VisualCR,
  title={Visual Commonsense R-CNN},
  author={Tan Wang and Jianqiang Huang and Hanwang Zhang and Qianru Sun},
  journal={2020 IEEE/CVF Conference on Computer Vision and Pattern Recognition (CVPR)},
  year={2020},
  pages={10757-10767}
}

@article{Yang2020DeconfoundedIC,
  title={Deconfounded image captioning: A causal retrospect},
  author={Yang, Xu and Zhang, Hanwang and Cai, Jianfei},
  journal={IEEE Transactions on Pattern Analysis and Machine Intelligence},
  volume={45},
  number={11},
  pages={12996--13010},
  year={2021},
  publisher={IEEE}
}

@article{Zhang2020CausalIF,
  title={Causal intervention for weakly-supervised semantic segmentation},
  author={Zhang, Dong and Zhang, Hanwang and Tang, Jinhui and Hua, Xian-Sheng and Sun, Qianru},
  journal={Advances in Neural Information Processing Systems},
  volume={33},
  pages={655--666},
  year={2020}
}

@article{Li2021CausalHM,
  title={Causal Hidden Markov Model for Time Series Disease Forecasting},
  author={Jing Li and Botong Wu and Xinwei Sun and Yizhou Wang},
  journal={2021 IEEE/CVF Conference on Computer Vision and Pattern Recognition (CVPR)},
  year={2021},
  pages={12100-12109}
}

@article{Yue2020InterventionalFL,
  title={Interventional few-shot learning},
  author={Yue, Zhongqi and Zhang, Hanwang and Sun, Qianru and Hua, Xian-Sheng},
  journal={Advances in neural information processing systems},
  volume={33},
  pages={2734--2746},
  year={2020}
}

@inproceedings{Nie2023ChestXI,
  title={Chest X-ray Image Classification: A Causal Perspective},
  author={Nie Weizhi et al.},
  booktitle={International Conference on Medical Image Computing and Computer-Assisted Intervention},
  pages={25--35},
  year={2023},
  organization={Springer}
}

@inproceedings{Nie2023InstrumentalVL,
  title={Instrumental Variable Learning for Chest X-ray Classification},
  author={Nie, Weizhi and Zhang, Chen and Song, Dan and Bai, Yunpeng and Xie, Keliang and Liu, Anan},
  booktitle={2023 IEEE International Conference on Systems, Man, and Cybernetics (SMC)},
  pages={4506--4512},
  year={2023},
  organization={IEEE}
}

@inproceedings{pmlr-v37-xuc15,
  title={Show, attend and tell: Neural image caption generation with visual attention},
  author={Xu Kelvin et al.},
  booktitle={International conference on machine learning},
  pages={2048--2057},
  year={2015},
  organization={PMLR}
}

@article{chestcon1,
  title={Training deep learning algorithms with weakly labeled pneumonia chest X-ray data for COVID-19 detection},
  author={Rajaraman, Sivaramakrishnan and Antani, Sameer},
  journal={MedRxiv},
  year={2020},
  publisher={Cold Spring Harbor Laboratory Preprints}
}

@inproceedings{chestcon2,
  title={Chestx-ray8: Hospital-scale chest x-ray database and benchmarks on weakly-supervised classification and localization of common thorax diseases},
  author={Wang, Xiaosong and Peng, Yifan and Lu, Le and Lu, Zhiyong and Bagheri, Mohammadhadi and Summers, Ronald M},
  booktitle={Proceedings of the IEEE conference on computer vision and pattern recognition},
  pages={2097--2106},
  year={2017}
}

@article{chestcon3,
  title={Deep long-tailed learning: A survey},
  author={Zhang, Yifan and Kang, Bingyi and Hooi, Bryan and Yan, Shuicheng and Feng, Jiashi},
  journal={IEEE Transactions on Pattern Analysis and Machine Intelligence},
  volume={45},
  number={9},
  pages={10795--10816},
  year={2023},
  publisher={IEEE}
}

@article{Liu2022CrossModalCR,
  title={Cross-Modal Causal Relational Reasoning for Event-Level Visual Question Answering},
  author={Yang Liu and Guanbin Li and Liang Lin},
  journal={IEEE Transactions on Pattern Analysis and Machine Intelligence},
  year={2022},
  volume={45},
  pages={11624-11641}
}

@inproceedings{M2I2,
  title={Self-supervised vision-language pretraining for medial visual question answering},
  author={Li, Pengfei and Liu, Gang and Tan, Lin and Liao, Jinying and Zhong, Shenjun},
  booktitle={2023 IEEE 20th International Symposium on Biomedical Imaging (ISBI)},
  pages={1--5},
  year={2023},
  organization={IEEE}
}

@inproceedings{Sonsbeek2023OpenEndedMV,
  title={Open-ended medical visual question answering through prefix tuning of language models},
  author={Van Sonsbeek, Tom and Derakhshani, Mohammad Mahdi and Najdenkoska, Ivona and Snoek, Cees GM and Worring, Marcel},
  booktitle={International Conference on Medical Image Computing and Computer-Assisted Intervention},
  pages={726--736},
  year={2023},
  organization={Springer}
}

@inproceedings{yang2021causal,
  title={Causal attention for vision-language tasks},
  author={Yang, Xu and Zhang, Hanwang and Qi, Guojun and Cai, Jianfei},
  booktitle={Proceedings of the IEEE/CVF conference on computer vision and pattern recognition},
  pages={9847--9857},
  year={2021}
}

\ifCLASSOPTIONcaptionsoff
  \newpage
\fi

\end{document}